\documentclass[letterpaper, 10 pt, journal, twoside]{ieeetran}   % Comment this line out if you need a4paper

\IEEEoverridecommandlockouts                              % This command is only needed if 
\usepackage{graphics} % for pdf, bitmapped graphics files
\usepackage{times}
\usepackage{cite}
\usepackage{multicol}
\usepackage[table]{xcolor}  
\usepackage{xcolor}
\usepackage[colorlinks,bookmarksopen,bookmarksnumbered,linkcolor=magenta,citecolor=green,urlcolor=teal]{hyperref}
\usepackage{amsmath}
\usepackage{amssymb}
\usepackage{algorithm}% http://ctan.org/pkg/algorithm
\usepackage{algpseudocode}% http://ctan.org/pkg/algorithmicx
\usepackage[font=footnotesize,labelfont=bf]{caption} % changing the caption size after reviewer 1 (assoc. ed.) comment
\usepackage{subcaption}
\usepackage{float}
\usepackage{pdfpages}
\usepackage{cleveref}
\newcommand{\shrink}{\def\baselinestretch{0.95}\large\normalsize}
\newcommand\footnoteref[1]{\protected@xdef\@thefnmark{\ref{#1}}\@footnotemark}

\title{\LARGE \bf
Using Collocated Vision and Tactile Sensors for Visual Servoing and Localization
}

\begin{document}

\title{Using Collocated Vision and Tactile Sensors for Visual Servoing and Localization
}
% \author{^{1}$, ,   and % <-this % stops a space
\author{Arkadeep Narayan Chaudhury$^{1}$, Timothy Man$^{1}$, Wenzhen Yuan$^{1}$ and Christopher G. Atkeson$^{1}$%
\thanks{Manuscript received: 9-9-21; Revised: 12-9-21; Accepted: 1-4-22}% <-this % stops a space
\thanks{This paper was recommended for publication by  Associate Editor J. Falco and
Editor Dan Popa upon evaluation of the Reviewers' comments.
This work was supported in part by the National Science Foundation through award 1717066 and the Toyota Research Institute.} %Use only for final RAL version
\thanks{$^{1}$All authors are with the Robotics Institute, Carnegie Mellon University, 
5000 Forbes Avenue, Pittsburgh, Pennsylvania, USA ZIP: 15232. Email addresses:
{ \{arkadeepnc, yuanwz, cga\}@cmu.edu}}%
\thanks{Digital Object Identifier (DOI): 10.1109/LRA.2022.3146565}
\thanks{Note: This archival version of the manuscript is significantly different in content from the reviewed and published version. The published version can be accessed here: https://ieeexplore.ieee.org/document/9699405.}
}
\markboth{IEEE Robotics and Automation Letters. Preprint Version. Accepted January, 2022}
{Chaudhury\MakeLowercase{\textit{et al.}}: Collocating vision and touch for localization } 
% \shrink
\maketitle

\shrink
\maketitle
\thispagestyle{empty}
\pagestyle{empty}

%%%%%%%%%%%%%%%%%%%%%%%%%%%%%%%%%%%%%%%%%%%%%%%%%%%%%%%%%%%%%%%%%%%%%%%%%%%%%%%%
\begin{abstract}
  Coordinating proximity and tactile imaging by collocating cameras with tactile sensors can 1) provide useful information before contact such as object pose estimates and visually servo a robot
  to a target with reduced occlusion and higher resolution compared to head-mounted or external depth cameras,
   2) simplify the contact point and pose estimation problems and help tactile sensing avoid erroneous matches when
  a surface does not have significant texture or has repetitive texture with many possible matches, and 3) use tactile imaging to further refine contact point and object pose estimation.
 We demonstrate our results  with objects that have more surface texture than most objects in standard manipulation datasets. 
We learn that optic flow needs to be integrated over a substantial amount of camera travel to be useful in predicting movement direction. Most importantly, we also learn that state of the art vision algorithms
do not do a good job localizing tactile images
on object models, unless a reasonable prior can be provided from collocated cameras. \newline
\end{abstract}

%%%%%%%%%%%%%%%%%%%%%%%%%%%%%%%%%%%%%%%%%%%%%%%%%%%%%%%%%%%%%%%%%%%%%%%%%%%%%%%%
\section{INTRODUCTION}
 \begin{figure}
    \centering
    \begin{subfigure}[b]{0.22\textwidth}
        \centering
        \includegraphics[width=1.1\textwidth]{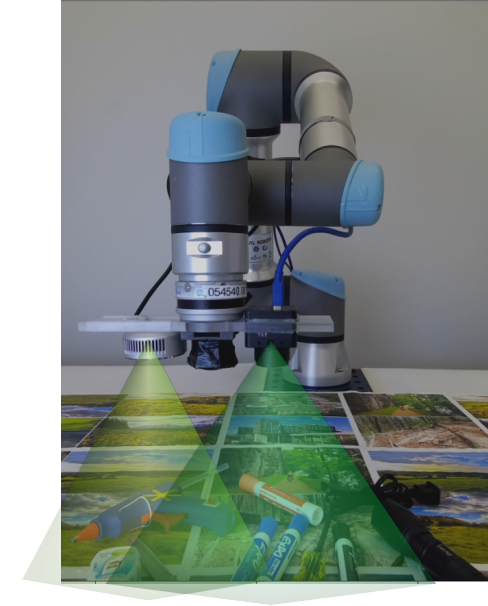}
        \subcaption{}
        \label{fig:cam_fovs}
    \end{subfigure}
    \hfill
    \begin{subfigure}[b]{0.22\textwidth}
        \centering
        \includegraphics[width=1.0\textwidth]{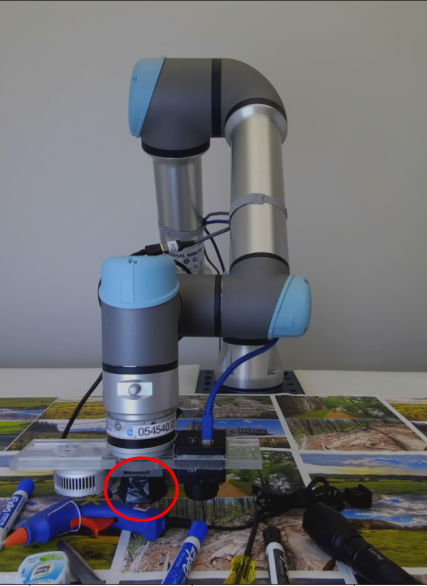}
         \subcaption{}
         \label{fig:gs_fovs}
    \end{subfigure}
    \\[ 2 ex] 
    \begin{subfigure}[b]{0.115\textwidth}
      \centering
      \includegraphics[width=\textwidth]{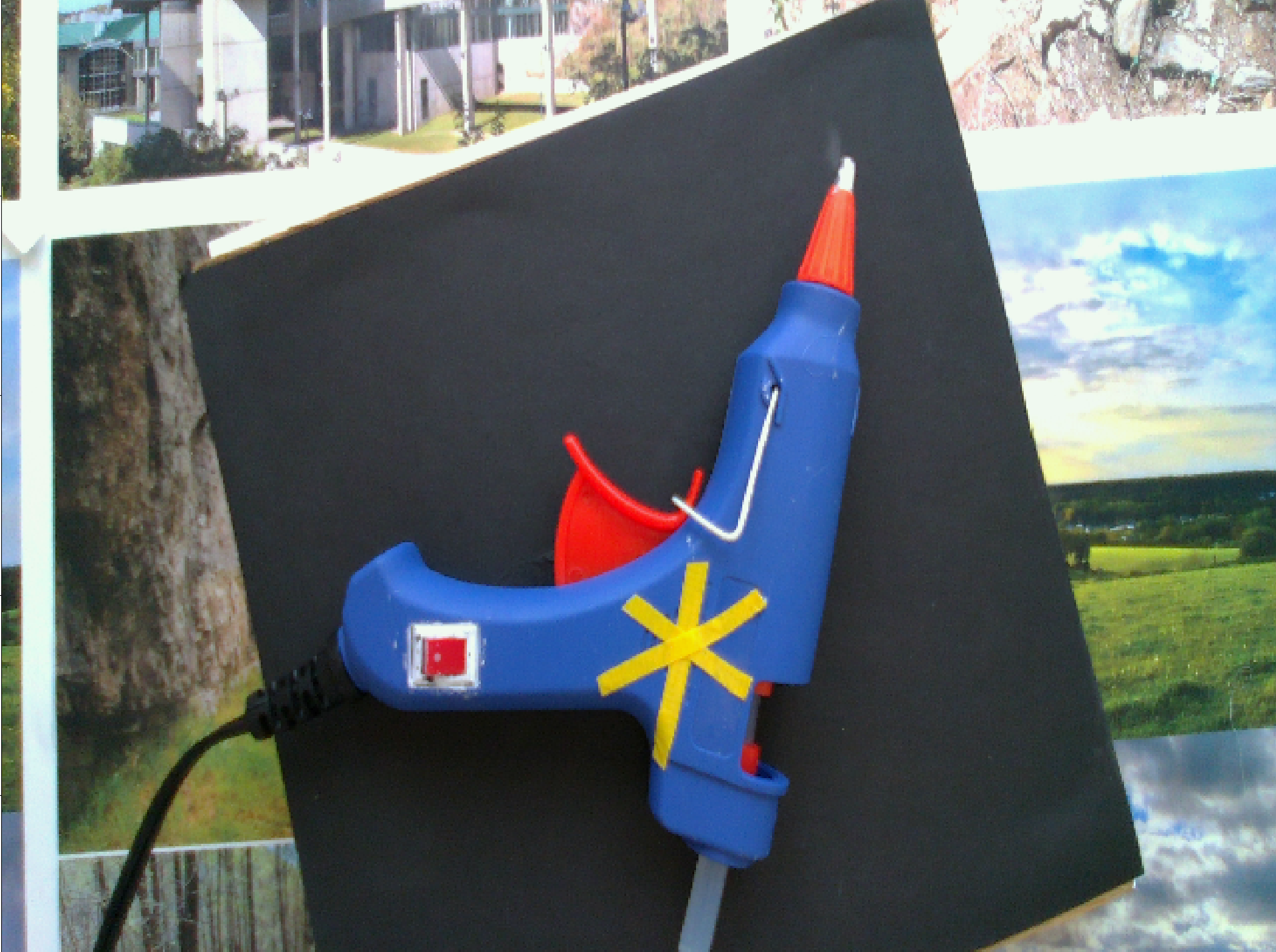}
      \caption{}
      \label{fig:cam_view}
    \end{subfigure} 
    \begin{subfigure}[b]{0.115\textwidth}
      \centering
      \includegraphics[width=\textwidth]{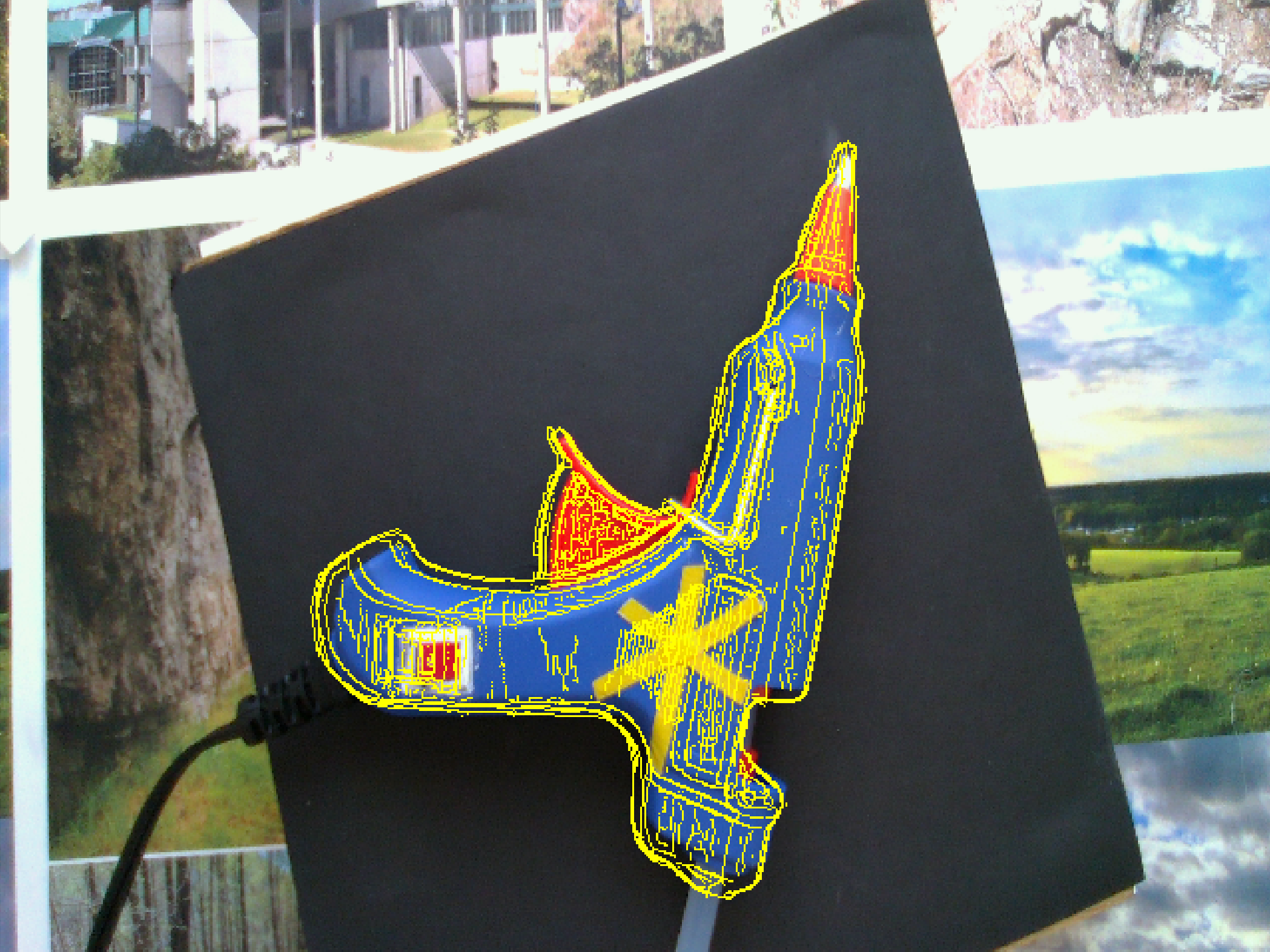}
      \caption{}
      \label{fig:cam_pose_est}
    \end{subfigure} 
    \begin{subfigure}[b]{0.115\textwidth}
      \centering
      \includegraphics[width=\textwidth]{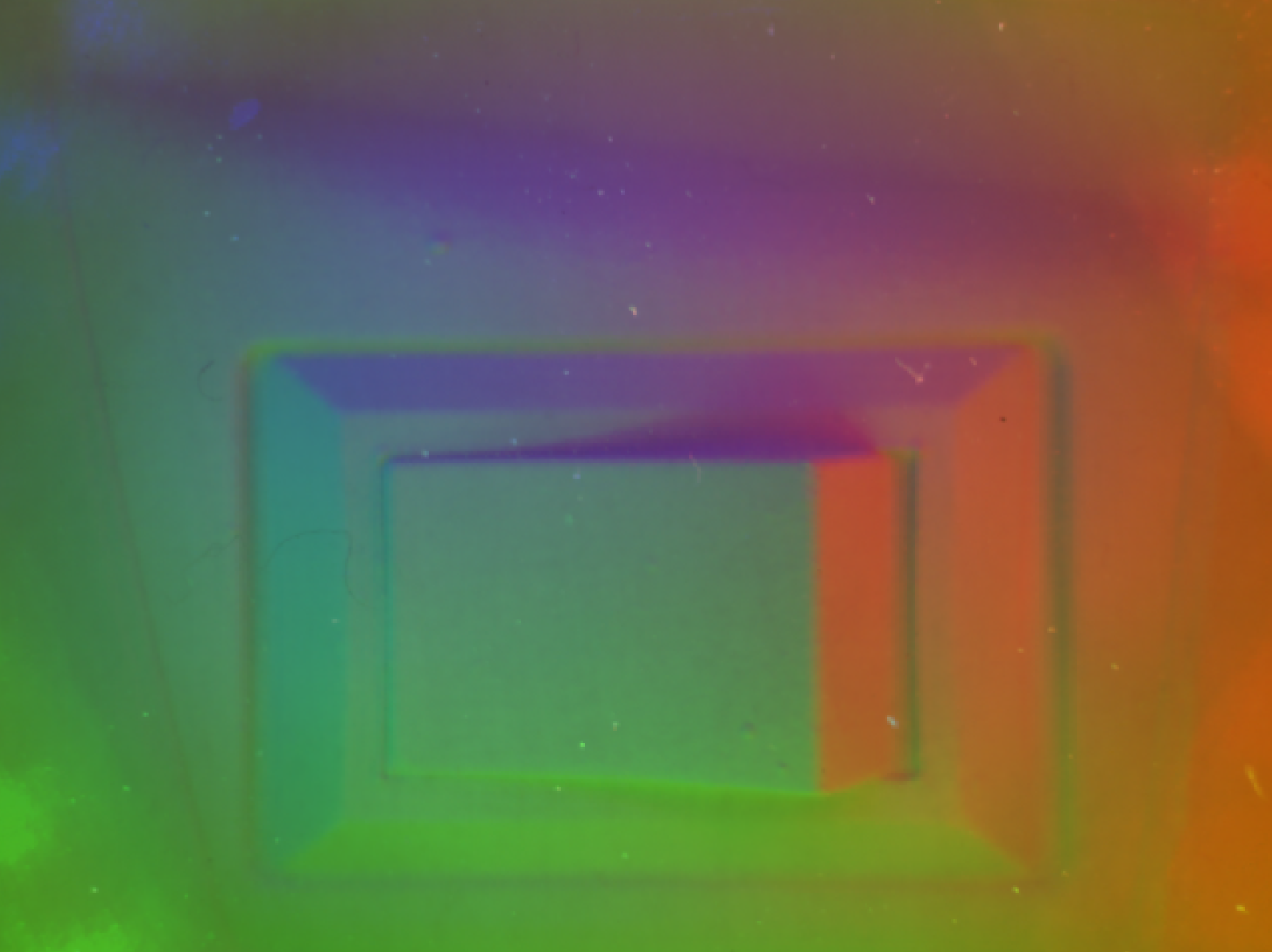}
      \caption{}
      \label{fig:gelsight_raw_data_opening_fig}
    \end{subfigure} 
    \begin{subfigure}[b]{0.115\textwidth}
      \centering
      \includegraphics[width=0.8\textwidth]{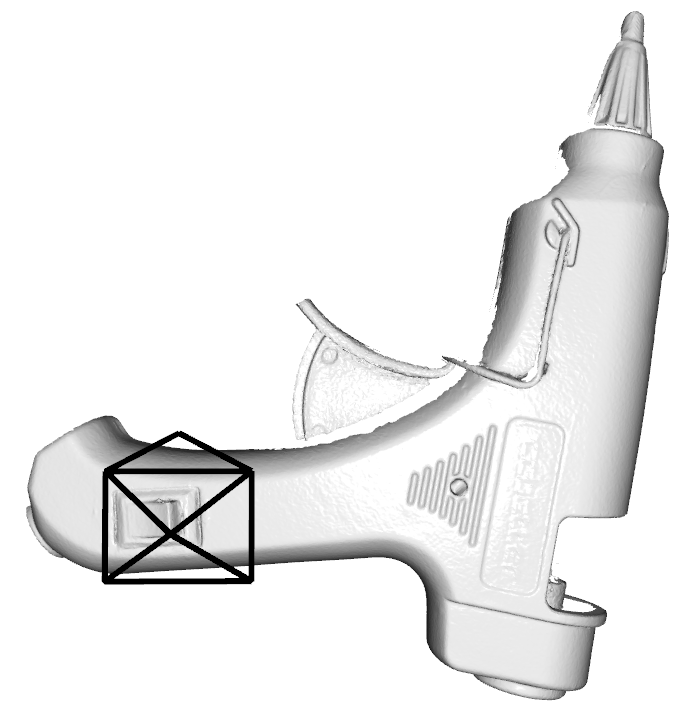}
      \caption{}
      \label{fig:gelsight_pose_est}
    \end{subfigure} 
       \caption{ We demonstrate an approach to integrate
      sensors with different fields of view to visually servo the robot arm to a predetermined contact point and estimate the pose of a fixed object relative to the sensors at contact. \Cref{fig:cam_fovs,fig:gs_fovs} show our sensor platform. We use 2 cameras with $70^\circ$ and $100^\circ$ fields of view, which we collocate with a modified GelSight sensor in the middle (\cref{fig:gs_fovs}). The cameras are used to visually servo the robot  (\cref{fig:cam_view})  and  generate  a preliminary pose estimate (\cref{fig:cam_pose_est}) while the robot is moving towards the target. At contact, the GelSight data is observed (\cref{fig:gelsight_raw_data_opening_fig}) and the preliminary pose is then refined to generate the object pose at contact. \Cref{fig:gelsight_pose_est} shows the camera pose superimposed on the mesh model of the object.}
       \label{fig:opening_figures}
   \end{figure}
  
 This  work  is  motivated  by FingerVision~\cite{Yamaguchi2016} where  the same  camera  was  used  for  tactile  sensing  and to  view nearby  objects  through  a  transparent  elastomer. Although FingerVision convinced us of the importance of proximity imaging, it did  not  provide high resolution images of contact surface texture,  and  the  transparent  elastomer  blurred proximity imaging, attracted  dust, and  got  scratched  and  worn so  the  view  of  nearby  objects was often not as good as we would like.  Separating  tactile and  proximity  imaging  enables  us  to  get  the  high  resolution of  GelSight~\cite{Yuan2017} tactile sensors  that produce images of the surface texture (tactile imaging), and  better  proximity  imaging  with  rigid lenses that don't attract dust as much, are easier to clean and don't scratch or get worn as easily.  This  paper  explores an   alternative to using the same camera for tactile and proximity imaging,  where a tactile sensor is collocated with a camera for proximity sensing (\cref{fig:opening_figures}). Since the tactile sensor we use, a GelSight variant, is also based on a camera, we are actually collocating multiple cameras to provide tactile and proximity imaging.

Cameras that move with a robot hand can have less occlusion and more resolution since they are closer to manipulated objects. Direct measurement of the direction or bearing to an object and its pose relative to the hand can be used to guide the hand to a particular contact location and center the hand with respect to an object. External and head-mounted cameras are often occluded by the robot itself as well as manipulated objects, and need to use stereo, multi-view, or other forms of depth measurement to locate the hand relative to the approach axis and object, which involves subtracting two noisy estimates (typically large numbers) to estimate a smaller quantity, which is usually less accurate than directly measuring the smaller quantity. We have found depth measurements from stereo or time of flight (TOF) cameras usually   have low spatial resolution, so getting the camera close to the hand is useful to improve depth resolution.

We divide the  problem of contact pose estimation into two parts – The initial phase before contact, when cameras can be used for vision-based servoing to a contact point target as well as estimating a prior for the tactile sensor contact point and object pose estimation, and the contact phase which refines the prior pose estimates. 
In this paper we assume that 1) the object is fixed, even during contact, 2) the object is a single rigid body with no articulations, and 3) we have a prior  3D model of the object (potentially provided by our vision of the current object) so we can express the pose of the object with respect to this model. For this paper we put aside the gross object localization and recognition problems in order to focus on fine localization, so we assume a vision system has already located the object, created a bounding box, and recognized the object by creating or selecting an appropriate 3D model that we want to register the actual object to (e.g. \cite{facebookresearch}). Our experimental pipeline involves selecting a goal and then visually servoing to that goal, recording color and depth data from the vision sensors, generating and maintaining pose estimates of the object, and using the estimates along with  tactile information received at contact to localize the contact point on the object. Through this work we show that: 
\begin{itemize}
 \item The optic flow, as observed by the hand mounted cameras, can be used to predict the heading direction of the hand using image-based techniques rather than
 3D geometry. In our setup frame-to-frame optic flow was dominated by small changes in orientation of the hand and thus the cameras. Optic flow had to be integrated across about 10cm of hand travel to be useful.
%  Optic flow from shorter movements was not reliable in predicting contact points due to small physical rotations of the camera caused both by interpolating the robot inverse kinematic solutions along the trajectory as well as unmeasured motions of the hand relative to the wrist joint angle sensors such as gear backlash and play.
\item A few ``mid-course" corrections can correct almost all the error in trajectories.  
\item  Pose errors, when measured only with the cameras  are  about $\pm1.5$ cm and  $\pm2^\circ$ in translation and rotation respectively about a vertical axis. 
\item Given these priors, tactile estimation based on a GelSight sensor
further improved the pose estimates to an uncertainty of $\pm1.5$mm
and $\pm0.5^\circ$ in translation and rotation respectively in cases distinct tactile signals were available.
% Without these priors it was difficult to localize the image from the GelSight using state of the art vision localization techniques.}
\item Collocated vision is particularly useful when an object does not
have distinctive tactile surface texture, or has repetitive surface texture. We show that using tactile sensing collocated with vision can help disambiguate tactile signals when used for localization. 
\end{itemize}
We provide additional details of our work, tables of results and reference implementations of our algorithms described in the paper here: \begin{footnotesize} \href{https://arkadeepnc.github.io/projects/collocated_vision_touch/index.html}{https://arkadeepnc.github.io/projects/collocated{\textunderscore}vision{\textunderscore}touch/index.html}
\end{footnotesize}

\section{Related work}\label{sc:Related_work}
In this section we provide a survey of related work on visual servoing and pose estimation using hand-mounted cameras and tactile sensing.
{\bf Recent work} on addressing these issues has used hand-mounted cameras to demonstrate superior performance in classical manipulation tasks such as grasping and bin picking\cite{Song2020}.
With the availability of a visual perspective  complementary to external (or head mounted) cameras, researchers have diversified the moving cameras to serve as tactile devices (\cite{Yamaguchi2016, Yuan2017}) and have implemented delicate manipulation behaviors (see e.g. \cite{Yamaguchi2017,Yuan2016}). Recent research has also developed tactile sensors and algorithms for estimating contact pose and inferring object from contacts \cite{Wang2018,Smith2020}, tracking object motion by fusing externally mounted cameras and tactile sensors\cite{Izatt2017}, and transferring information between external cameras and hand-mounted cameras (see e.g. \cite{Li2019}). A closely related work by \cite{Luo2015_contact_localizaton} discusses  integration of a visual and tactile measurement through a Bayesian filter. 
  
{\bf Vision-based localization and contact prediction:}\label{sc:rel_work_in_hand_cam}
Camera-on-hand or more generally camera-on-mobile-agent arrangements have been investigated by several researchers to pursue diverse goals such as visual servoing to a workspace goal (e.g. \cite{Kelly2000}), collision avoidance systems on miniature aerial vehicles (e.g. \cite{Mori2013}), and how flying insects, birds,
and rapidly moving animals perceive motion\cite{Serres2017}.
Literature on quantitative analysis of looming\footnote{Looming or visual looming is defined as the phenomena of an object getting bigger in the visual field as the relative distance between the observer and the object decreases.} (e.g. \cite{Raviv1992}, \cite{Raviv2000}) is of particular interest to us as we try to identify an area in the image space corresponding to the direction of heading of the robot at any particular moment. Recent research on optical expansion (e.g. \cite{Yang2020}) is focussed on supervised learning, instead of hand crafted functions (e.g. \cite{Raviv1992, Raviv2000}) to compute dense scene flow from optic flow to identify relative motion of objects and the agents, and has been demonstrated to exhibit state of the art performance in identifying objects heading towards the agent. In the current work we build upon research on optic flow for scene understanding to identify an area in the robot's visual field corresponding to the physical point in the workspace where the robot is currently headed. 

 \begin{figure*}[t]
    \centering
    \begin{subfigure}[b]{0.35\textwidth}
        \centering
        \includegraphics[width=1.0\textwidth]{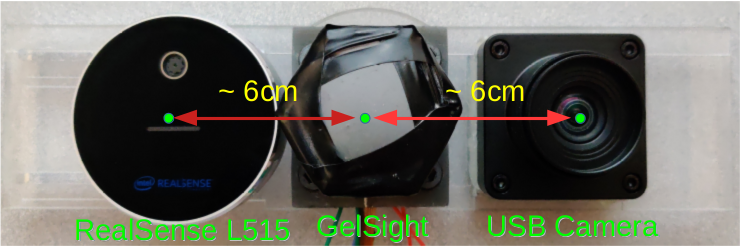}
        \subcaption{}
        \label{fig:new_sensor}
    \end{subfigure}
    % \hfill
    \begin{subfigure}[b]{0.20\textwidth}
        \centering
        \includegraphics[width=1.02\textwidth]{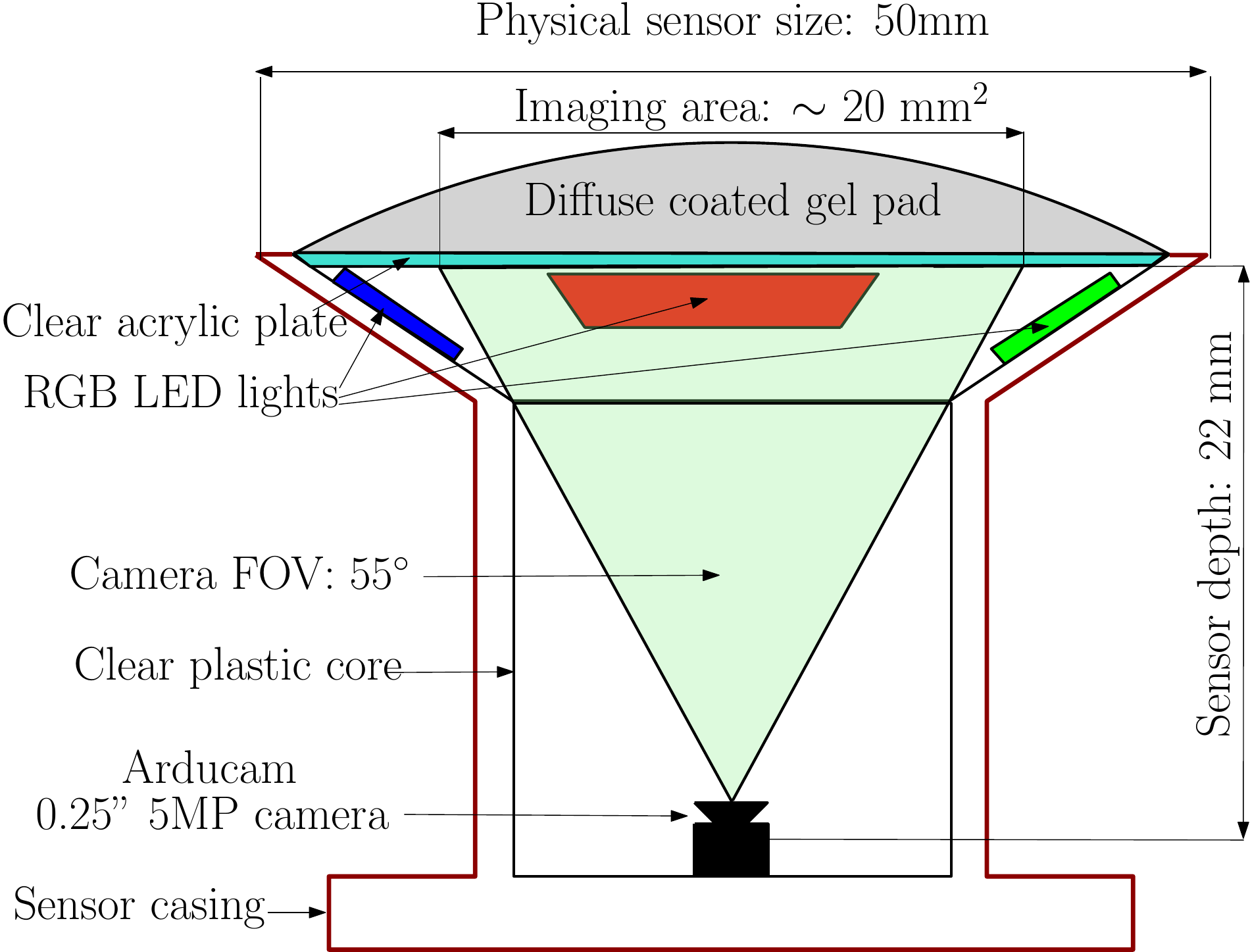}
         \subcaption{}
         \label{fig:gelsight_cross_sn}
    \end{subfigure}
    \begin{subfigure}[b]{0.20\textwidth}
      \centering
      \includegraphics[width=1.02\textwidth]{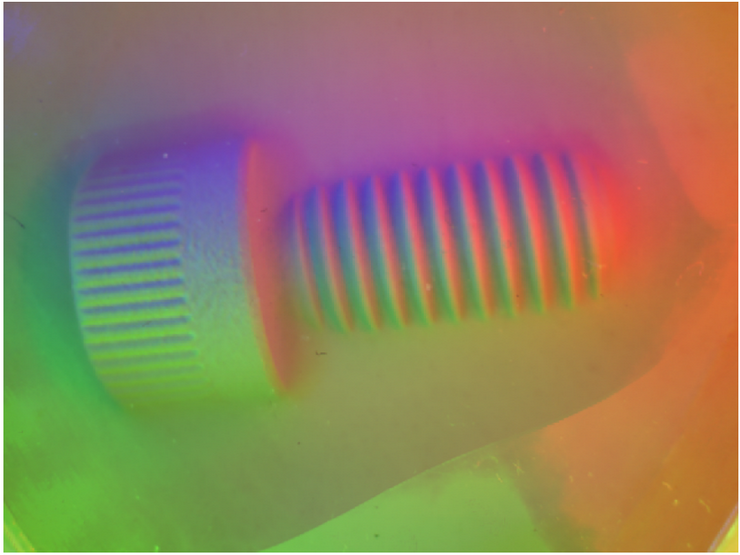}
       \subcaption{}
       \label{fig:gelsight_raw_data}
  \end{subfigure}
  \begin{subfigure}[b]{0.20\textwidth}
    \centering
    \includegraphics[width= 1\textwidth]{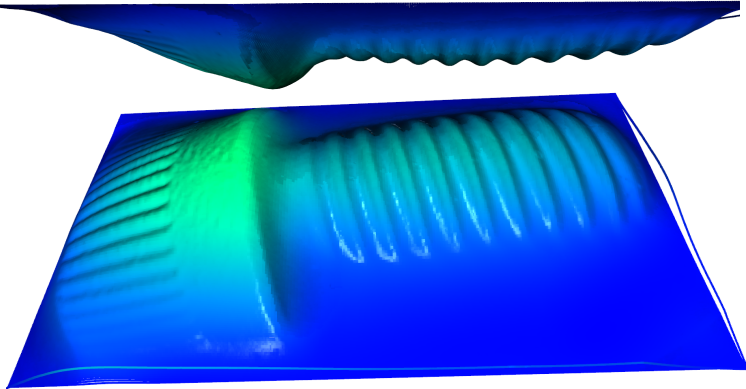}
     \subcaption{}
     \label{fig:Gelsight_proc_data}
\end{subfigure}
\caption{\Cref{fig:new_sensor} shows our sensor platform which is attached to the robot manipulator. The sensors, from left to right, are the Intel RealSense L515 LiDAR camera, our modified version of the GelSight and an RGB camera. The RGB camera and the RealSense are at a distance of 6 cm each from the camera to the left and right respectively. \Cref{fig:gelsight_cross_sn} shows the schematic of our modified version of the GelSight, \cref{fig:gelsight_raw_data} is an instance of the raw data collected by our GelSight when pressed against an 18mm long 6mm diameter bolt with 1mm pitch. Through the image, we note the physical scope of the sensor is almost entirely covered by the 18mm long bolt. We process the raw GelSight data to yield metric depth and normal maps. The depth of the reconstructed surface is rendered as a shaded point cloud in 2 views in \cref{fig:Gelsight_proc_data}.}
\label{fig:sensor_platform}
\end{figure*}

%%%%%%%%%%%%%%%%%%%%%%%%%%%
{\bf Tactile localization and contact estimation:}\label{sc:rel_work_tactile_loc}
% Tactile sensing is a relatively newer area of perception research which has gained traction mainly because a tactile sensor measurement is complementary to a vision sensor measurement in the context of perception. 
 Vision sensors can have a wide ``field of view'' and are good for making large scale
models. A tactile sensor has a much smaller measurement area (or field of view) and can potentially capture minute details of the surface it interacts with. Early research on tactile sensing leveraged this capability, even with a seemingly low resolution tactile sensor (Weiss Robotics DSA9205), to demonstrate object recognition using image feature descriptors\cite{Luo2015} and, recognize and localize an articulated object through a sequence of touches (see e.g. \cite{Luo2016}). With the introduction of camera-based higher resolution tactile sensors, most notably the GelSight (see \cite{Yuan2017}) and its derivative GelSlim\cite{Donlon2018}, investigations on tactile object recognition and localization have made significant progress in tactile sensing driven perception. \cite{Li2014} described tactile localization using the GelSight sensor using conventional feature based image alignment. \cite{She2019} integrated the GelSlim with a gripper and interfaced it with a model based controller to successfully perform re-grasps of a cable. More recently, \cite{Bauza2019} demonstrated tactile localization and shape reconstruction using a GelSlim sensor, where the authors trained 
neural networks to generate height maps with ground truth data obtained from robot experiments with the sensor and known objects. The trained network was then used to generate  height maps of the surface of an object and the height maps were registered to reconstruct the object surface. This work was extended by \cite{Bauza2020} where the authors used a renderer to generate and cache a large number of possible tactile signals of objects from a data set touching a tactile sensor (GelSlim in this case) in different orientations. An actual tactile signal corresponding to a particular object at a particular pose was then compared with the cache to retrieve candidate object and pose pairs and the network demonstrated in \cite{Bauza2019} was then used to generate a height map which was registered with the candidate object at the candidate pose to localize contact. In \cite{Sodhi2020} the authors describe a method to integrate a learned representation of a tactile signals (from a GelSight in this case) as a factor in a task of estimating object pose with a stream of subsequent measurements.
\cite{Alspach2019} introduced a new time of flight based tactile sensor, the soft-bubble, and demonstrate object identification using learned embeddings and object localization by registering the tactile signal (obtained as a point cloud by their sensor) with the retrieved geometric model of the recognized object. Recently, and perhaps closest to the current work, \cite{Suresh2021} introduced a method to reconstruct the shape of an object by refining a coarse prior shape obtained by an RGBD sensor using tactile measurements from a GelSight attached to a robot. Also related to the current work, Dikhale et al.\cite{Dikhale2022} demonstrated a method to fuse off-the-shelf learned pose estimators (PoseCNN \cite{Xiang2017} in this case) from external RGBD sensors and in-hand tactile sensor outputs to estimate the poses of grasped objects.  In the current work we build upon literature on contact localization using high resolution tactile sensors, visual localization and hand mounted cameras to demonstrate a suite of collocated vision based sensors that can be used to visually servo a robot to touch and localize objects in the workspace.

\section{Methods}\label{sc:methods}
In this section we describe our sensor platform (A) and 
algorithms to estimate where the robot hand will go (B1),
visually servo the robot to a target contact point (B2),
estimate the pose of the target object (C), and combine
vision and tactile information to estimate both the contact
point and refine the object pose estimate (D).
 
\subsection{Sensor platform}\label{sc:sensor_platform}
In this work, we collocate cameras with a camera-based tactile sensor by putting the cameras and the tactile sensor in close physical proximity while operating them independently. The sensor platform consists of a GelSight tactile sensor in the middle (co-incident with the robot wrist's axis) (\cref{fig:opening_figures,fig:sensor_platform}), with a camera on either side. We modify the GelSight as described by \cite{Johnson2011} in the physical sensor form factor introduced by \cite{Yuan2017}. We use a LIDAR-based RGBD sensor (Intel RealSense L515) with a $70^\circ$ field of view to provide depth (\cref{fig:cam_fovs,fig:new_sensor} left) and color images of the workspace, and a USB camera (a Sony IMX 291 sensor) with a wider $100^\circ$ field of view lens (\cref{fig:cam_fovs,fig:new_sensor} right).
% \textcolor{blue}{We chose these cameras partly because they were of comparable size to our GelSight sensor. This criterion eliminates other popular RGBD cameras based on stereo vision, such as various versions of the Kinect} 
We chose these cameras partly because they were of comparable size to our GelSight sensor, thus ruling out most other popular RGBD cameras. Also, the USB camera yielded high resolution images which were better than the color channel of our 3D sensor and helped us localize small objects.
% Also, a time-of-flight (TOF/LIDAR) camera still works well when there are few visual features.}
Further details on the sensors can be found in \cref{fig:sensor_platform}.

\begin{figure}[!]
    \centering
    \begin{subfigure}[b]{0.22\textwidth}
        \centering
        \includegraphics[width=0.65\textwidth]{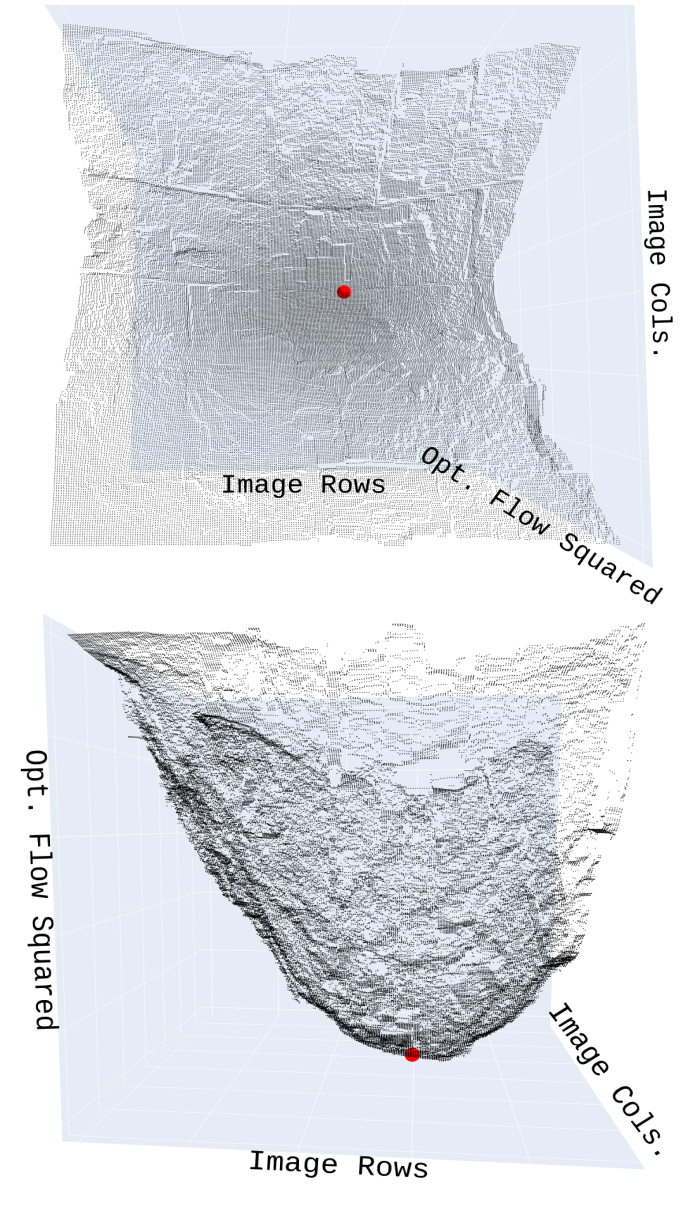}
        \subcaption{}
        \label{fig:opt_flow_quad_surfaces}
    \end{subfigure}
    % \hfill
    \begin{subfigure}[b]{0.22\textwidth}
        \centering
        \includegraphics[width=1.01\textwidth]{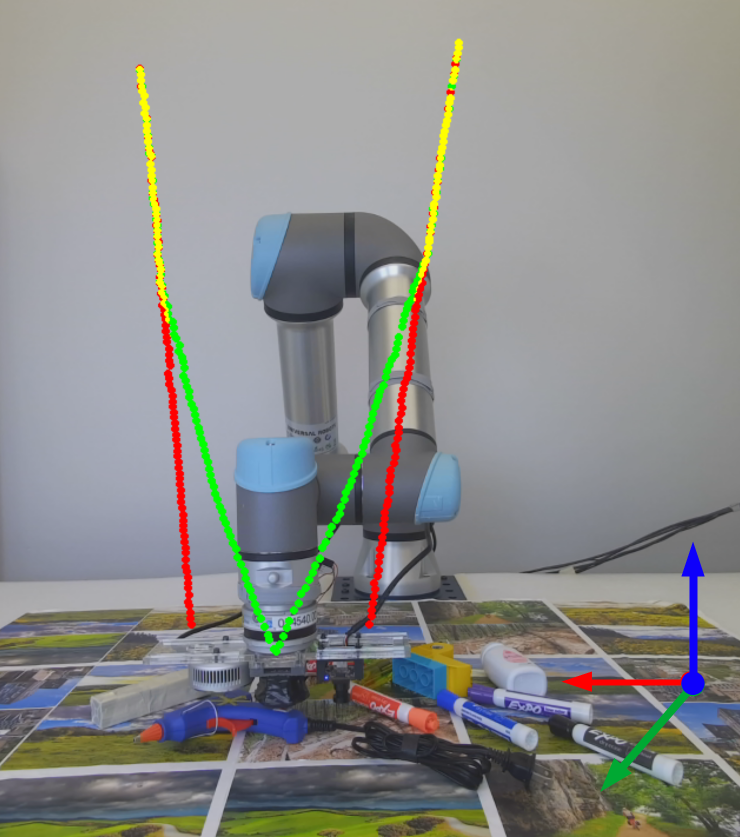}
         \subcaption{}
         \label{fig:traj_corrections}
    \end{subfigure}
       \caption{\Cref{fig:opt_flow_quad_surfaces} shows the surface of the squared magnitude of the optical flow between a consecutive frame pair in 2 views. We note that this surface assumes a parabolic shape. The red dot is the minima of the optical flow surface as identified by our algorithm. \Cref{fig:traj_corrections} demonstrates the usage of our algorithm to correct trajectory errors using both cameras as shown in \cref{fig:new_sensor}. The X, Y and Z axes are marked in red, green and blue in \cref{fig:traj_corrections} on the bottom right of the figure}
       \label{fig:opt_flow_traj_corr}
\end{figure} 

\subsection{Visual servoing with hand mounted cameras}\label{sc:visual_servoing}
In this section we describe a method to estimate the robot hand's heading direction in the workspace and then we describe how to servo to a goal using that estimate. 
\subsubsection{Optical point of expansion from in hand cameras}\label{sc:POE_description}
\begin{figure*}
    \centering
     \begin{subfigure}[b]{0.31\textwidth}
      \centering
      \includegraphics[width=\textwidth]{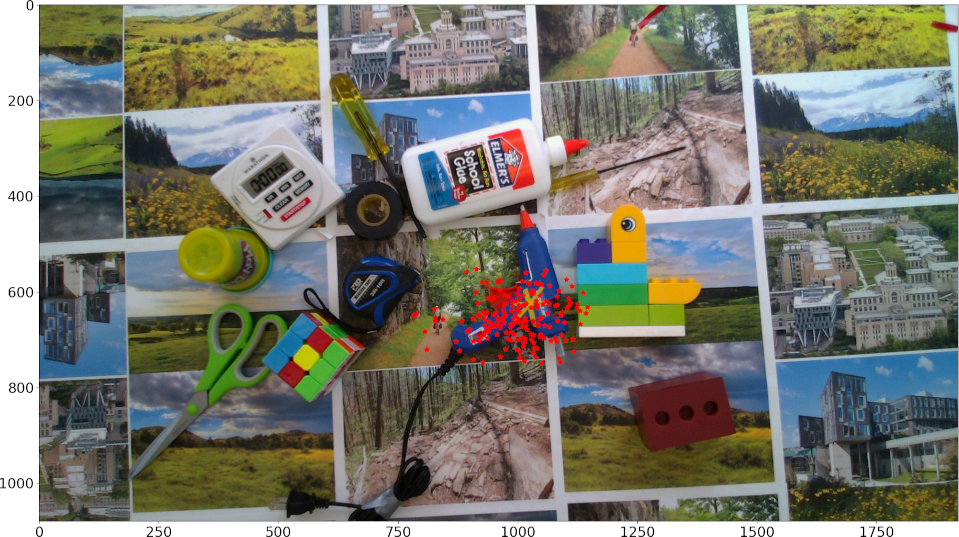}
      \subcaption{Int.: 1cm; St.dev.: (284,98)}
      \label{fig:1cm}
  \end{subfigure}
       \begin{subfigure}[b]{0.31\textwidth}
      \centering
      \includegraphics[width=\textwidth]{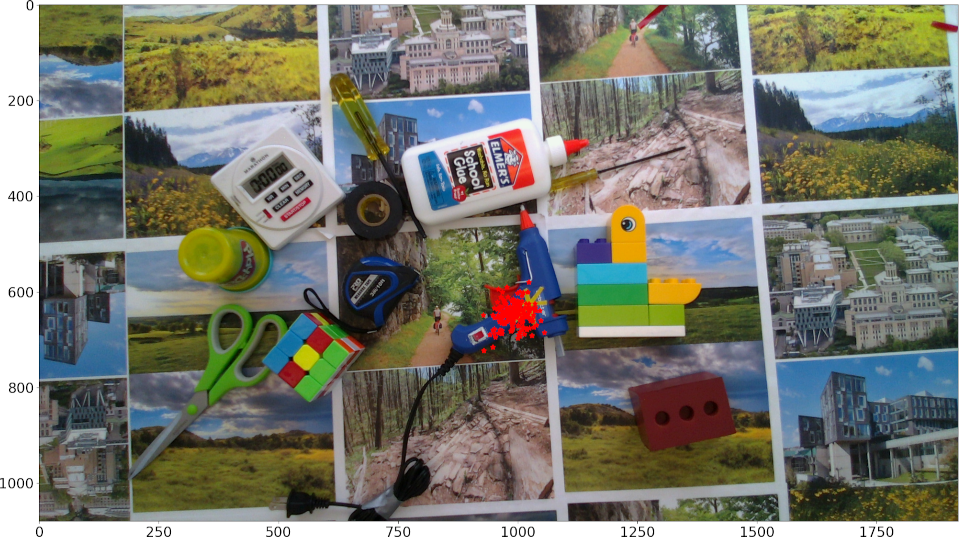}
      \subcaption{Int.: 5cm; St.dev.: (146,60)}
      \label{fig:5cm}
  \end{subfigure}
       \begin{subfigure}[b]{0.31\textwidth}
      \centering
      \includegraphics[width=\textwidth]{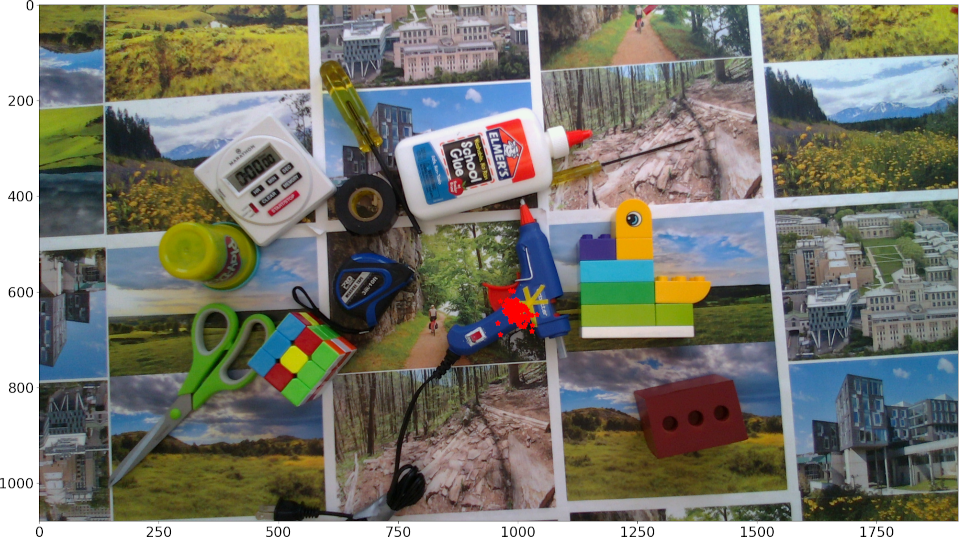}
      \subcaption{Int.: 7.5cm; St.dev.: (62,55)}
      \label{fig:7_pt_5cm}
  \end{subfigure}
     \begin{subfigure}[b]{0.31\textwidth}
      \centering
      \includegraphics[width=\textwidth]{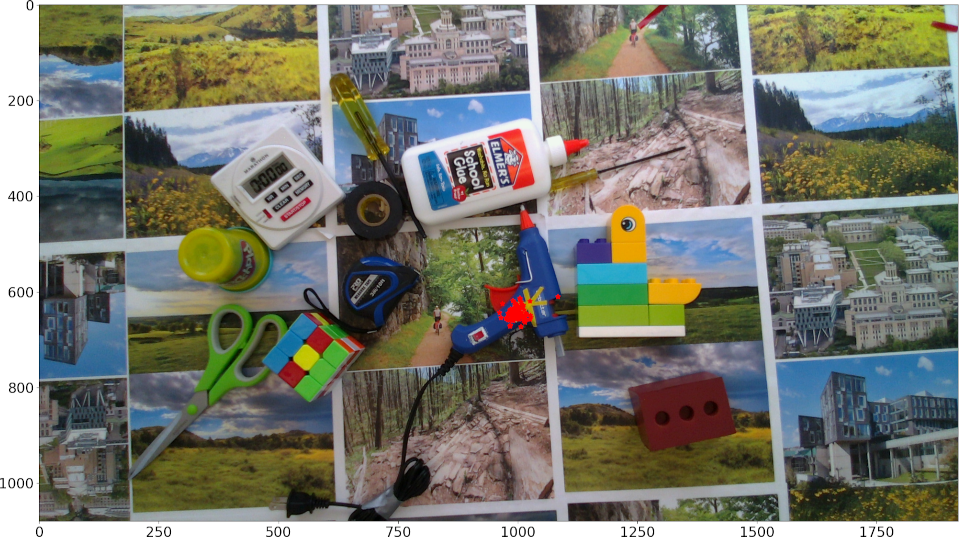}
      \subcaption{Int.: 10cm; St.dev.: (19,27)}
      \label{fig:10cm}
  \end{subfigure}
       \begin{subfigure}[b]{0.31\textwidth}
      \centering
      \includegraphics[width=\textwidth]{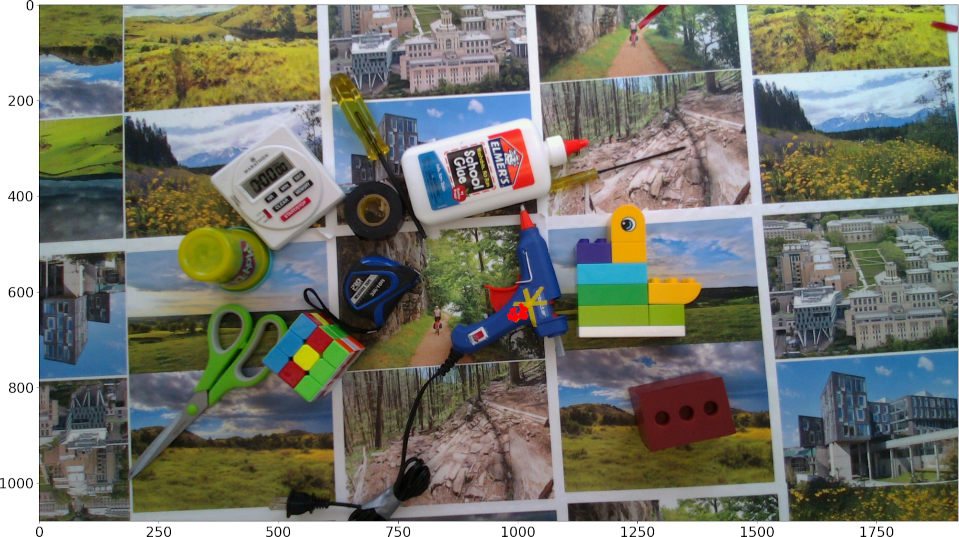}
      \subcaption{Int.: 15cm; St.dev.: (21,20)}
      \label{fig:15cm}
  \end{subfigure}
       \begin{subfigure}[b]{0.31\textwidth}
      \centering
      \includegraphics[width=\textwidth]{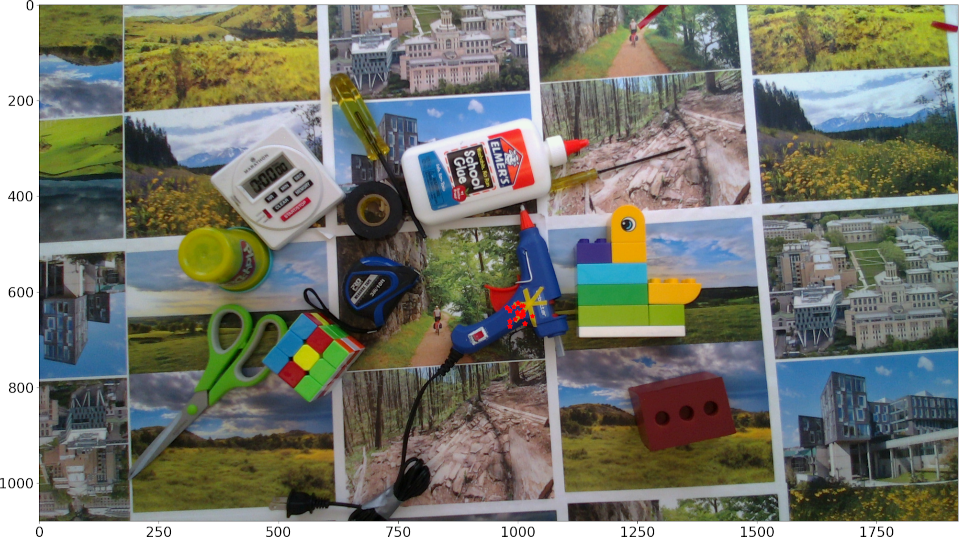}
      \subcaption{Int.: 20cm; St.dev.: (17,15)}
      \label{fig:20cm}
  \end{subfigure}
    \caption{In this experiment, we move the robot vertically down by 65 cm to a goal location slightly below the yellow cross mark on the handle of the glue gun. There are no errors in the goal location being tracked for this case. This experiment is repeated 10 times. The red dots are the predictions of potential point of contact (calculated as the instantaneous POE). We note that the predictions are centered about the actual point of contact -- a point slightly below the yellow cross mark on the glue gun. We report 6 cases where we predict the  potential point of contact by looking at various intervals of the trajectory. We report the interval lengths (in cm) and the standard deviation in predicting the point of contact (in pixels) as the labels of the figures. We note that as we increase the length of the interval, the standard deviation of the prediction decreases (as seen through ``clumping" of the predicted potential points of contact), but the number of possible predictions decreases (as seen through fewer number of red dots with increasing interval sizes). This leads us to conclude that the averages across larger temporal (and spatial) windows produce smoother and more stable error signals (or correction signals in the case of visual servoing). For our use case, averaging across 10 cm intervals provided us with ``enough'' number of correction signals while having reasonably low variance. }
    \label{fig:poe_all}
\end{figure*}

We process the scene to identify the location of a 3D point corresponding to the heading direction of the robot in the image space. To achieve this, we calculate the optical flow between the consecutive frames obtained by the hand mounted cameras and identify the region in the image from which the optic flow seems to be emerging (i.e., we look for a portion of the scene which has zero translation) as the camera moves towards the scene. Assuming that the world scene is relatively flat (object depth $\ll$ projection depth), the square of the magnitude of the optic flow at each pixel is roughly distributed as a parabolic surface (see \cref{fig:opt_flow_quad_surfaces}). We calculate the motion field (per pixel optic flow magnitude and direction) between  two consecutive frames using the OpenCV implementation\cite{OpenCV2021} of the Lucas-Kanade dense optical flow, which solves for per pixel motions (along horizontal and vertical directions) over the full image. We also tested the Farenb{\"a}ck optical flow \cite{Farnebaeck2003} and the Brox optical flow \cite{Brox2004}, and found that the dense Lucas-Kanade optical flow performs slightly better in computation speed and produces smoother optic flow fields.
The optical point of expansion (POE) is obtained as the minima of the surface representing the square of the magnitude of the optic flow. We use a robust algorithm described in \cref{alg:POE_calculation} to detect the POE. 
\begin{algorithm}
  \caption{Calculating the point of expansion}
  \label{alg:POE_calculation}
  \begin{algorithmic}[1]
    \Procedure{calculatePOE }{image stream $\{I\}$}
    \State Make pointers list $\{L\}$ for overlapping image tiles 
    \For{$(I_t,~I_{t-1})\in I$}
    \State $\{d_x, d_y\}\longleftarrow$ denseLKOpticalFlow($I_t,~I_{t-1}$) 
    \State $d_t\longleftarrow {d_x^2+d_y^2}$, $\theta_t\longleftarrow\arctan(\dfrac{d_y}{d_x})$
    \State $d_t\longleftarrow$ normalizeAndHistogramEqualize($d_t$) 
    \For{\texttt{tile $\in L$}} \Comment{\boxed{\emph{exec. with n workers}}}
      \If{Histogram($\theta_t$[\texttt{tile}]) is uniform}
      \State \texttt{tile}$\rightarrow L_{ang}$
      \EndIf
      \State add one random tile to $L_{ang}$ \Comment{if tracking bad }
    \EndFor
    \For{\texttt{tile $\in L_{ang}$}} \Comment{\boxed{\emph{exec. with n workers}}}
      \State \texttt{fit} $\leftarrow$ Least sq. fit paraboloid to $d_t$[\texttt{tile}] 
    \EndFor
    \State choose $k$ best fits from \texttt{fit} $\rightarrow$ \texttt{best fits}
    \State find pixel $P$ of minima of the \texttt{fit} $\in$ \texttt{best fits}
    \State POE $\leftarrow$ $P$ with the most uniform Histogram around $\theta_t$[P])
    \EndFor
    \State \textbf{return} list of POEs per frame pair
    \EndProcedure
  \end{algorithmic}
\end{algorithm}
\newline \indent
The heading direction estimated by the optical flow between consecutive frames was too noisy to yield meaningful heading estimates. To address this, we looked at the trajectory correction estimates for each of the on hand cameras and found their prediction to be very closely correlated (almost equally incorrect or equally correct), which led us to rule out camera noise and incorrect robot kinematics including the camera mounts as the cause of the noise. We concluded that the errors were being caused by small unmeasured rotations of the robot wrists (play or backlash).
Numerical modeling showed that the expected amount of play in orientation led to optic flow values comparable to the ``noisy" shifts
in the POE. We also noted that the noise in the predicted trajectory errors was centered about zero. We averaged POE shifts over a robot travel of at least about 10 cm to get usable POE estimates. For our robot setup. where a maximum of 1m downward travel was possible, empirically we observed that about 10cm non-overlapping intervals provided useful trajectory corrections and provided the opportunity for several corrections as the robot moved to the target. We describe the experiment that led us to this conclusion in \cref{fig:poe_all}.
% averaging the trajectory ove r 10 cm was empirically chosen between averaging over 5cm, 15cm and 20cm, as it provided the maximum number of corrections possible during the robot's motion towards the goal, while yielding reasonable trajectory error estimates due to a larger smoothing window.}
\subsubsection{Correcting trajectory errors using pixel space errors}\label{sc:Goal_correction}
In the previous section we described a method to identify the image coordinates of the point in the workspace to which the robot is headed. In this section we address the problem of correcting trajectory errors using those predictions. This can be useful when the robot's trajectory needs correction and the only information about the updated goal is available in pixel space -- possibly from a hand mounted RGB camera which detected a movement of the target. We use the following intuition for visual servoing: to reach a pre-planned goal in the robot's workspace, the robot, almost always, needs to move towards that goal, hence the image of the goal point and the observed POE
should be very close. Errors in the pixel space between these two, indicates an error in the trajectory being followed by the robot, which we aim to correct. 
% To do this, we note that as the cameras are registered to the robot, we can compute the Jacobian which correlates the pixel space error to the task space error and we can invert the Jacobian (up to an arbitrary scale in the projection direction) to obtain trajectory space corrections from pixel space errors.
In more precise terms, let us denote the true workspace goal as $\mathbf{X}_G^W$ and it's estimate $\widehat{\mathbf{X}_G^W}$ in homogeneous 3 space. The robot is initially planned to move to $\widehat{\mathbf{X}_G^W}$ and let the POEs obtained for the subsequent frames be $\mathbf{p}$, in the pixel space. As the camera is registered to the robot, a given point in time, we can calculate the world to camera projection matrix as $\mathbf{P} = \mathbf{K}[\mathbf{R}_{world}^{cam}|\mathbf{t}_{world}^{cam}]$
and can project $\mathbf{X}_G^W$ and it's estimate $\widehat{\mathbf{X}_G^W}$ to $\mathbf{x}_G$ and $\widehat{\mathbf{x}_G}$ in homogeneous 2 space. If the motion between two frames captured by the camera-in-hand is mostly perpendicular to the imaging plane, we can approximate $\widehat{\mathbf{x}_G}$ with the point of expansion $\mathbf{p}$ for each subsequent frames. We also note that, as $\mathbf{p}$ is in the pixel space, with this approximation, we discard the ``z-buffer'' of the projective transformations associated with $\mathbf{x}_G$ and as a consequence, cannot correct for a error in the camera's projective axis (camera's z axis). However, this is not a problem for us as the location of the table is known, and we combine the correction signals from the 2 in hand cameras to mitigate possible errors due to rejecting the ``z-buffer''.  \newline
\indent We  use the camera projection Jacobian and calculate the task-space error as \cref{eq:trajectory_corretion}
\begin{align}\label{eq:trajectory_corretion}
  \Delta = \left[\begin{array}{ccc}
    \frac{f_x}{Z_W} & 0 & -\frac{X^W_G f_x+Z^W_G c_x}{{Z^W_G}^2}+\frac{c_x}{Z^W_G} \\
    0 & \frac{f_y}{Z^W_G} & -\frac{Y^W_G f_y+Z^W_G c_y}{{Z^W_G}^2}+\frac{y}{Z^W_G}
  \end{array} \right]^+ \left[\begin{array}{c}
  p^u -   \mathbf{x}_G^u   \\
  p^v - \mathbf{x}_G^v
  \end{array} \right]
\end{align}
In \cref{eq:trajectory_corretion}, $[f_x, f_y, c_x, c_y]$ are the camera intrinsics, and we denote $\mathbf{X}_G^W = [X^W_G, Y^W_G, Z^W_G]$ and the $[\cdot]^+$ denotes the Moore-Penrose pseudo-inverse. As an example use case, we use the following procedure to correct trajectory goals as the robot moves approximately 1m towards the true workspace goal $\mathbf{X}_G^W$ with a erroneous initial estimates of upto $\pm10$cm in each of X and Y directions. \newline
% we can extend the POE calculation discussed in the previous section to correct for robot trajectory errors in heading  using pixel errors between the projected pixel location of the workspace goal and the POE as observed by the hand mounted cameras.
% the procedure described below to correct for  an error of 10cm along both the X and Y directions on a trajectory of length 1m covered in 100 steps. 
\begin{itemize}
    \item steps 1--20:  Plan a path with 100 waypoints till goal, and execute the first 20 steps.
    \item steps 21--40: For the next 20 steps, calculate the POE and then use the camera intrinsics and pose of the camera with respect to the robot base to obtain the mean error in trajectory space. Correct the trajectory error and plan for the remaining 60 steps.
    \item steps 41--60: Execute the current path re-planned at step 41, do not track the point of expansion or generate new trajectory error estimates. 
    \item steps 61--80: Repeat steps 21--40 while moving through the corrected trajectory, at the end correct trajectory again and re-plan if necessary.
    \item steps 81--100: Continue with the last planned trajectory, and stop if goal is reached to given tolerance. 
  \end{itemize}
In \cref{fig:traj_corrections}, on the two sides of the robot, we show 2 trajectories where the robot travels about 1m (vertically) from the start to the final contact position, and each of the trajectories need a correction of 10cm errors in the X and Y directions in the robot workspace. We show the initial portions of the trajectories in yellow, the planned portions in red, and the corrected portions in green. The final accuracy achieved was within 5mm of the target. 
% \textcolor{blue}{We also note that we are correcting trajectory errors (in $\mathbb{R}^3$) using pixel space errors (in $\mathbb{R}^2$) as the robot is moving to its goal. This problem is not immediately solved using robot kinematics and accurate ground truth measurements of the goal error in robot workspace, which we do not have access to.}
    \begin{figure*}
    \begin{subfigure}[b]{0.15\textwidth}
        \centering
        \includegraphics[width=\textwidth]{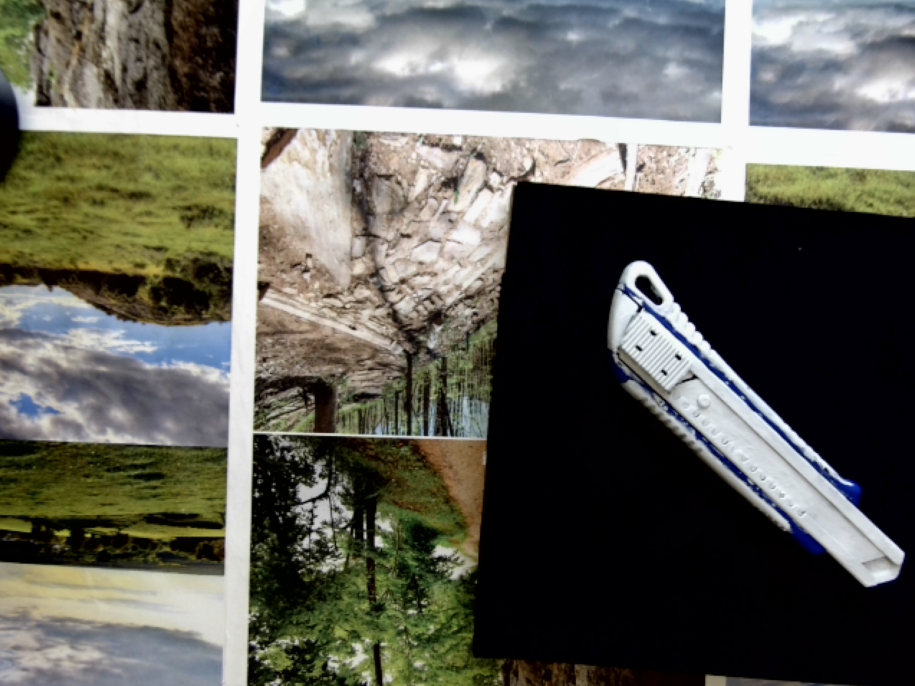}
        \caption{}
        \label{fig:pose_est_col_img}
    \end{subfigure}
   \begin{subfigure}[b]{0.15\textwidth}
    \centering
    \includegraphics[width=\textwidth]{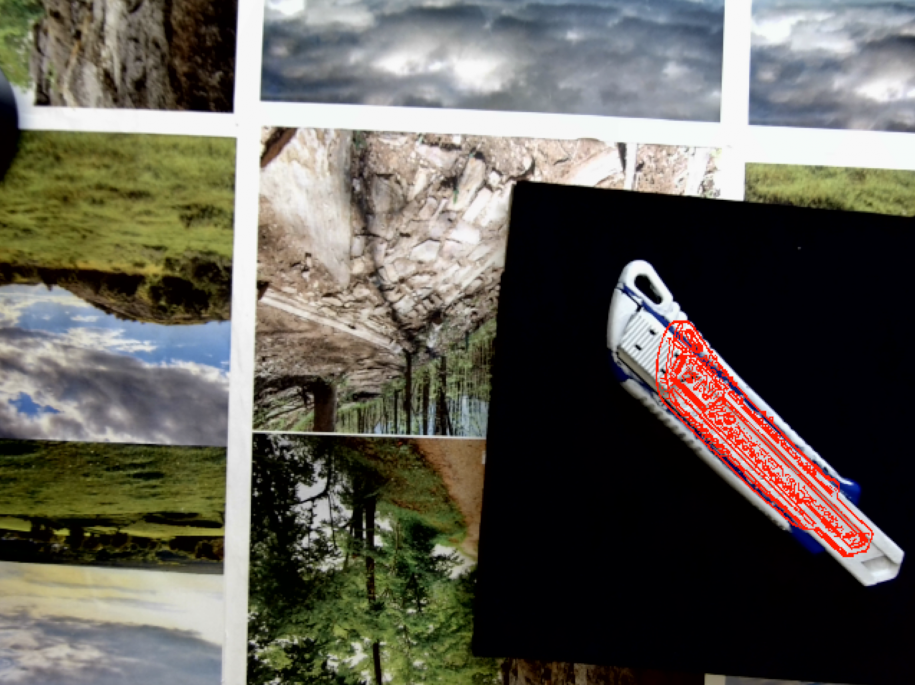}
    \caption{}
    \label{fig:pose_est_coarse_result}
    \end{subfigure}
    \begin{subfigure}[b]{0.15\textwidth}
    \centering
    \includegraphics[width=\textwidth]{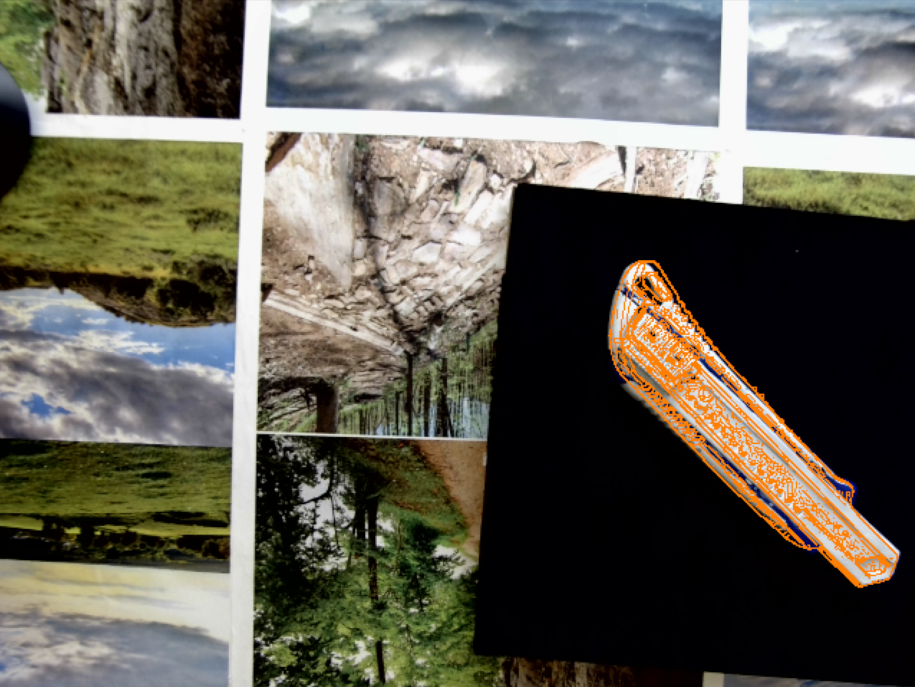}
    \caption{}
    \label{fig:pose_est_rgb_loc_result}
    \end{subfigure}
    \begin{subfigure}[b]{0.15\textwidth}
    \centering
    \includegraphics[width=\textwidth]{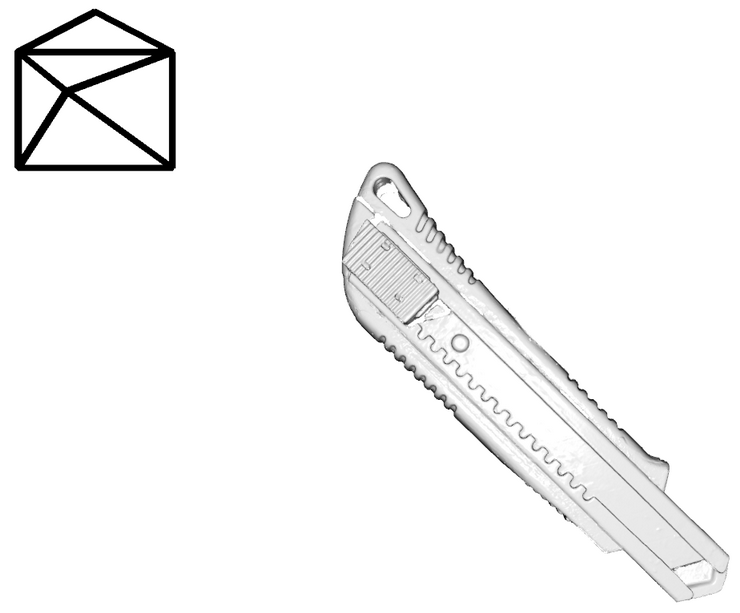}
    \caption{}
    \label{fig:pose_est_pose_transfer_gs}
    \end{subfigure} 
    \begin{subfigure}[b]{0.15\textwidth}
    \centering
    \includegraphics[width=\textwidth]{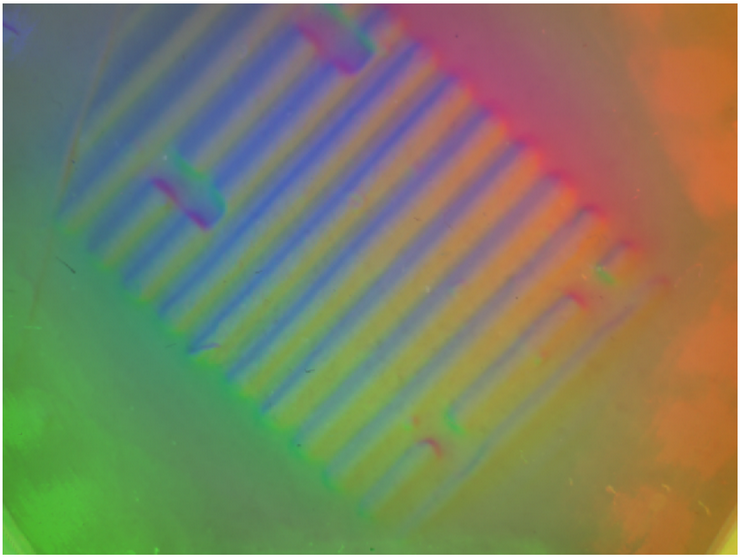}
    \caption{}
    \label{fig:pose_est_raw_gs_data}
    \end{subfigure}   
   \begin{subfigure}[b]{0.15\textwidth}
    \centering
    \includegraphics[width=0.8\textwidth]{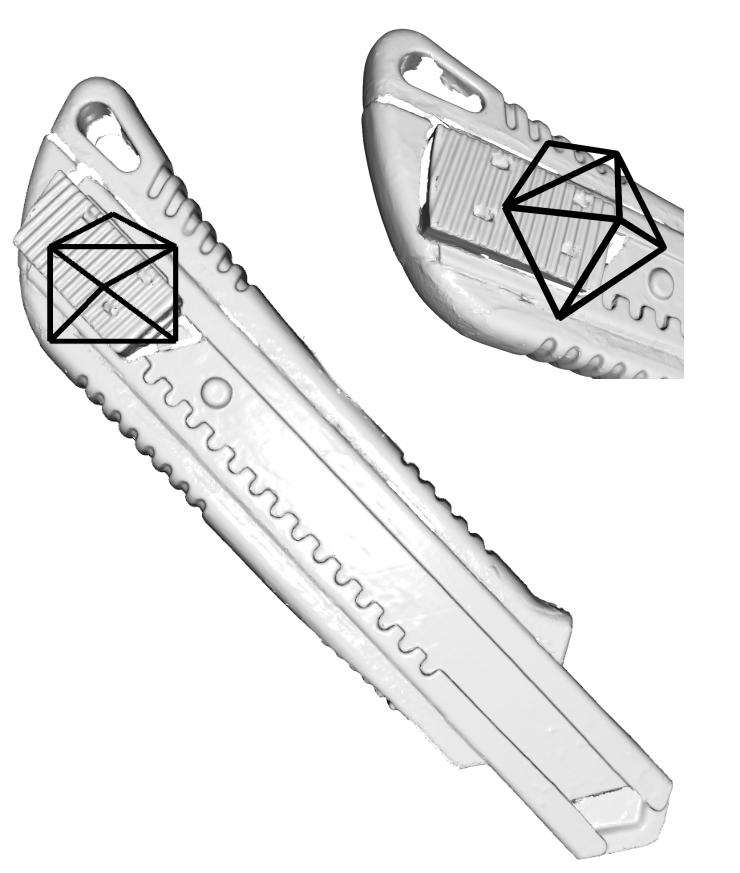}
    \caption{}
    \label{fig:pose_est_gs_final}
    \end{subfigure}
    \caption{Our pipeline to estimate pose through vision and touch. \Cref{fig:pose_est_col_img} is the color channel of the RGB-D sensor stream captured at a particular point in the trajectory. \Cref{fig:pose_est_coarse_result} is the pose estimate obtained after solving \cref{eq:sparse_edge_loss}, which is refined by solving \cref{eq:dense_edge_cost} and we obtain a camera pose estimate $\hat{\zeta}$ (shown in \cref{fig:pose_est_rgb_loc_result}) which is mostly correct in the camera projection depth and camera yaw. We transfer $\hat{\zeta}$ to the reference frame of the GelSight and obtain \cref{fig:pose_est_pose_transfer_gs} showing the relative pose of the object and the GelSight. This corresponds to the pose $\hat{\zeta}_{GS}$. This estimate is further refined by minimizing \cref{eq:gelsight_alignment} using the GelSight data (\cref{fig:pose_est_raw_gs_data}) obtained at contact, and we obtain the final pose shown in \cref{fig:pose_est_gs_final} }
    \label{fig:pose_estimation_master}
    \end{figure*}
\shrink
\subsection{Pose estimation through vision}\label{sc:visual_pose_estimation}
In this section we discuss visual localization of an object with known 3D geometry. At least in the early stages of approach, the hand-mounted cameras can see all or large portions of the target object, and standard image based registration methods (\cite{Liu2012, Imperoli2015}) can be used to estimate the object's pose relative to the robot hand. This is quite different from the situation with tactile sensing where only a small part of the object is visualized. As is common in the whole-object-visible camera-based pose estimation literature (see e.g. \cite{Liu2012,Imperoli2015} and \cite{Choi2012}), we decompose pose estimation into two parts -- coarse pose estimation by aligning the centroids and edge moments, and finer alignment using a distance-based cost applied densely.  \newline \indent As a prerequisite for this part we need the edge pixels of the object in the image and for this we use either the depth edges on the object if available or else use the image gradient edges of the object --  we use a Canny edge finder for this purpose. Let us denote this binary edge image as $\mathbf{I}_S$. Next, we formulate an optimization problem to identify an object pose that produces the most similar edge distribution to $\mathbf{I}_S$. \newline 
\indent To do this, we generate an initial guess for $\theta$ (the angle of rotation about the camera projection axis) for the object from the principal components of the camera image and if an aligned depth map is available, we use its mean to initialize the $z$ component (distance between the camera and object along the camera projection axis), or else we initialize the $z$ depth arbitrarily. The $x$ and $y$ directions along the image plane are initialized by back-projecting the centroid of the edge pixels using the camera intrinsics and the initial value of $z$. With the initial guess $\omega = [x,y,z,0,0,\theta]$, we use a differentiable renderer (we use a modified version of the DIRT renderer from \cite{Henderson2019}) to render the mesh model of the object, extract the corresponding edge image $\mathbf{I}_R(\omega)$ and solve the following minimization to obtain a rough pose estimate from the camera image. For each $\mathbf{I}_R$, we identify the image edge pixels $\mathbf{p}_R^i$ and $\mathbf{p}_S$ for $\mathbf{I}_R$ and $\mathbf{I}_S$ respectively and minimize the following sparse edge matching cost $\mathbb{E}_{cs}$ in \cref{eq:sparse_edge_loss}
\begin{align}\label{eq:sparse_edge_loss}
  \mathbb{E}_{cs}(\omega) &= \gamma \left[||\bar{\mathbf{p}_R}(\omega) - \bar{\mathbf{p}_S}||_2 \right] +(1-\gamma)\\ &\langle \mathcal{V}\left(\mathbf{p}_R(\omega) - \bar{\mathbf{p}_R} \right) \cdot \mathcal{V}\left(\mathbf{p}_S - \bar{\mathbf{p}_S} \right) \rangle \nonumber
\end{align}
where, $\mathcal{V}(\mathbf{p})$ is the direction of largest variance of the mean centered point set $\mathbf{p} \in \mathbb{R}^2$, given by the eigenvector corresponding to the maximum eigenvalue, $\langle \cdotp,\cdotp\rangle$ is the dot product between vectors and $\gamma$ is a weighting factor. Minimizing \cref{eq:sparse_edge_loss} aligns the centroid and approximately recovers the angle of rotation along the camera projection axis. \newline 
\indent The expression for $\mathbb{E}_{cs}$ does not admit automatic gradient calculation due to the non-differentiable selection of pixel indices to obtain $\mathbf{p}_R$ from $\mathbf{I}_R$, therefore, we obtain a finite difference gradient using central differences. We minimize $\mathbb{E}_{cs} \forall \theta_i$ and obtain the candidate pose parameters $\hat{\omega} = \{X_{cs}, Y_{cs}, Z_{cs}, \theta_{cs}\}$ in the camera coordinate frame, corresponding to the minimum $E_{cs}$. 
From \cref{fig:pose_est_coarse_result} we note that minimization of $\mathbb{E}_{cs}$ is not expected to solve for the projection depth, as it only recovers the orientation of the camera and not the projection depth. In the next part, we solve the projection depth by minimizing a modified version of the dense differentiable cost using the directional chamfer matching energy as discussed in \cite{Liu2012} and \cite{Imperoli2015} (\cref{eq:orientation_aware_chamfer_match_cost}). 
\begin{align}\label{eq:orientation_aware_chamfer_match_cost}
 \mathbb{E}_{cm}(\mathbf{I}_S, \mathbf{I}_R^\zeta) &= \sum_{^\xi\mathbf{p}_s^i\in ^\xi\mathbf{I}_S}\left[\min_{^\xi\mathbf{p}_R^j\in ^\xi\mathbf{I}_R(\zeta)} || ^\xi\mathbf{p}_S^i - ^\xi\mathbf{p}_R^j(\zeta)||\right] 
\end{align}
In \cref{eq:orientation_aware_chamfer_match_cost}, the outer sum $\sum_{^\xi\mathbf{p}_s^i\in ^\xi\mathbf{I}_S}(\cdot)$ implements the edge awareness by binning the edges according to their orientation (as quantized by $\xi \in [-\pi, \pi]$), and implicitly assigning edge pixel correspondences. We observe that the inner minimization problem $\min_{^\xi\mathbf{p}_R^j\in ^\xi\mathbf{I}_R(\zeta)} || ^\xi\mathbf{p}_S^i - ^\xi\mathbf{p}_R^j(\zeta)||$ for each $^\xi\mathbb{E}_{cm}$ can be solved by the Euclidean distance transform. As the $^\xi\mathbb{E}_{cm}$ cost is cumulative over the $^\xi\mathbf{I}_S$ image, the cost boils down to the pixel-wise sum of absolute differences between the Euclidean distance transforms (EDT) of images  $^\xi\mathbf{I}_S$ and  $^\xi\mathbf{I}_R(\zeta)$. So using the definition of Euclidean distance transform from \cite{Felzenszwalb2012} in \cref{eq:orientation_aware_chamfer_match_cost}, we simplify our dense edge matching energy as
\begin{align}\label{eq:dense_edge_cost}
  \mathbb{E}_{cm}(\zeta) &= \sum_{\xi} \left[ \sum_{\mathcal{G}}\left[|\rm{EDT}(^\xi\mathbf{I}_S) - \rm{EDT}(^\xi\mathbf{I}_{R}(\zeta))|\right] \right]
\end{align}
We minimize this function with gradient descent to obtain a coarse pose estimate $\hat{\zeta}$ from the camera image and transfer the pose estimate to the GelSight camera frame as $\hat{\zeta}_{GS}$. In contrast to \cite{Imperoli2015,Liu2012}, we implemented modified and differentiable versions of the matching costs and thus, our gradient steps are about 80\% faster than the reference implementations of \cite{Imperoli2015,Liu2012} for the same size of the image. In practice, to keep the computational cost low, we maintain a coarse pose estimate using \cref{eq:sparse_edge_loss} throughout the major part of the trajectory and switch to \cref{eq:dense_edge_cost} at a point beyond which the objects are de-focussed. For objects in \cref{fig:opening_figures,fig:pose_estimation_master,fig:pose_estimation_folding_knife} this distance was around $15$cm and for objects in \cref{fig:pose_estimation_key_chain_and_cam_ckt} it was $10$cm.
%%%%%%%%%%%%%%%%%%%%%%%%%%%
\shrink
\subsection{Precise contact pose estimation through touch}\label{sc:gelsight_alignment}
We found that our vision-based localization had errors on the order of a centimeter, due to our algorithms imperfect camera
calibration, local minimums in matching, and sometimes a lack of visual features to match or track. In addition,
as the hand comes near the object, the depth and image sensor measurements become unusable -- the LiDAR based sensor provides reliable depth estimates at distances greater than 25cm, and the camera image measurements were usable at distances greater than  12-15cm to the object, after which the items of interest often go out of view and the image becomes too blurry to extract high quality edges needed by our 3D pose estimation algorithms. We find that using the tactile image on contact can improve contact
point and pose estimation. In \cref{fig:sensor_platform} we noted that we could generate metrically correct depth and normal maps of the deformed GelSight surface at contact. In this section, we use that information, along with the pose estimated from the previous section to localize contact, given that we know the geometry of the object. To achieve this, we use multi-scale dense depth and normal map alignment to obtain the pose of object with respect to the tactile sensor.\newline
\indent We capture the tactile image in our GelSight's native camera resolution of $(640\times480)$  pixels and obtain depth and normal maps. We then decompose each of the normal and depth maps into 4 lower pyramid levels. We denote these 5 normal and depth maps of the source image as $N_S$ and $D_S$ respectively. Next, we render the object (using \cite{Henderson2019}) through the GelSight's viewport using the pose estimated in the previous section and obtain normal and depth maps corresponding to the frame sizes of $N_S$ and $D_S$ respectively. \newline 
\indent We note here that this is not an exact simulation of the GelSight sensor through our renderer -- the ideal GelSight should only measure objects touching it i.e. $\sim$25mm from the camera. Anything beyond or closer than that is either not touching the sensor or interfering with it. We relax this requirement, and also neglect the effects of the soft body contact between the gel and the object to get a better basin of convergence in the optimization problem. We denote these sets of rendered depth maps by $N_R(\hat{\zeta})$ and $D_R(\zeta)$. Our alignment cost function $\mathbb{E}_{gs}(\zeta)$ for the GelSight data is 
\begin{align}\label{eq:gelsight_alignment}
  \mathbb{E}_{gs}(\zeta) = \sum_{i=1}^5\left[\left|D_S^i - D_R^i(\zeta)\right| + \left[1-\langle N_S^i \cdot N_R^i (\zeta) \rangle\right] \right]
\end{align}
where $\langle N_S^i \cdot N_R^i \rangle$ denotes the pixel-wise dot products of the normal maps. We solve this with $\hat{\zeta}_{GS}$ as the starting guess using gradient descent. \Cref{fig:pose_estimation_master} describes the steps discussed above. 
\section{Results} \label{sc:results}
In this section we present the results of localizing contacts using the sensor setup and the algorithms discussed in this work through 6 scenarios. For the experiments presented below, we use a black background to simplify the object segmentation and edge detection. Except for the glue gun, all the other objects had reflective parts and were painted matte white to remove specularities. 
%In case of the glue gun, we had to choose the power switch because that was the most feature rich part of the object.}
\subsubsection{Localization with color and depth edges}\label{sc:loc_col_and_depth}
\begin{figure}[!]
  \begin{subfigure}[b]{0.15\textwidth}
      \centering
      \includegraphics[width=\textwidth]{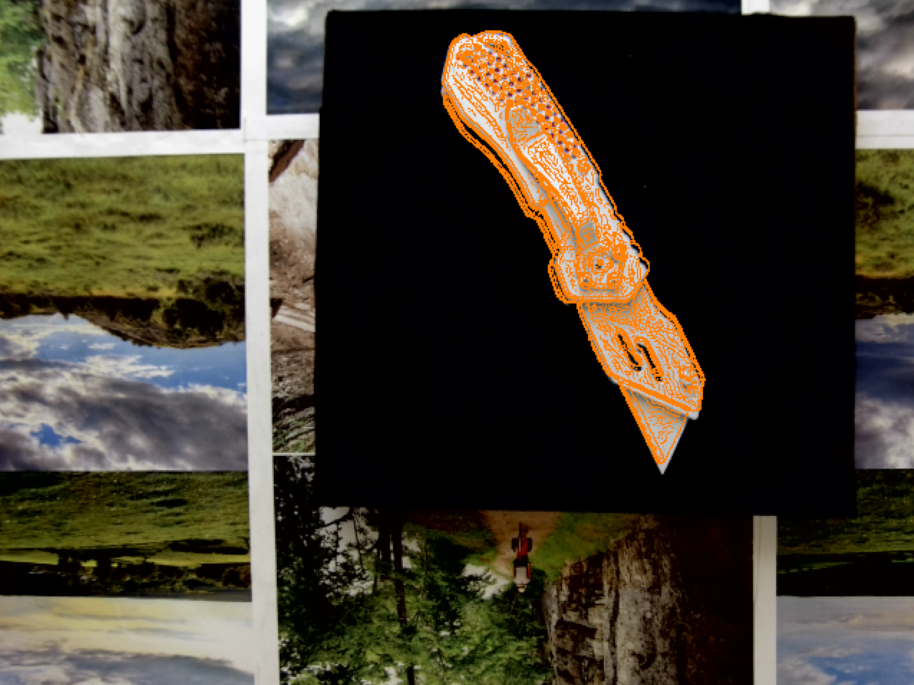}
      \caption{}
      \label{fig:pose_est_rgb_folding_knife}
  \end{subfigure}
 \begin{subfigure}[b]{0.15\textwidth}
  \centering
  \includegraphics[width=\textwidth]{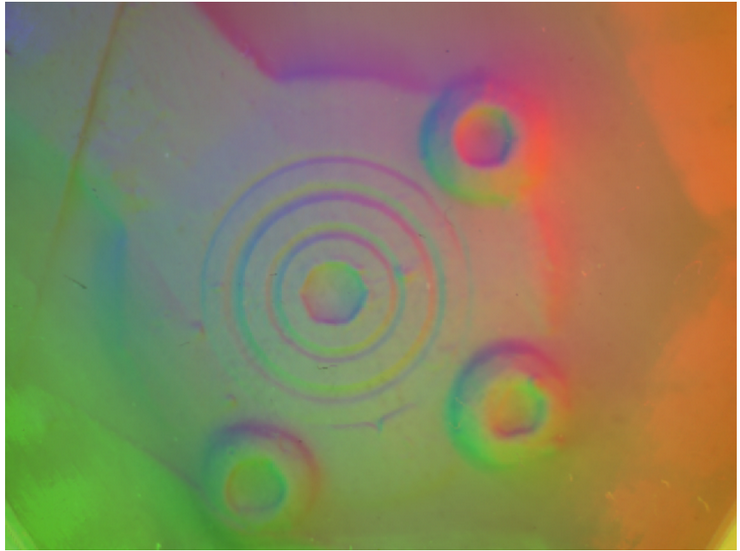}
  \caption{}
  \label{fig:pose_est_gs_raw_folding_knife}
  \end{subfigure}
  \begin{subfigure}[b]{0.15\textwidth}
  \centering
  \includegraphics[width=\textwidth]{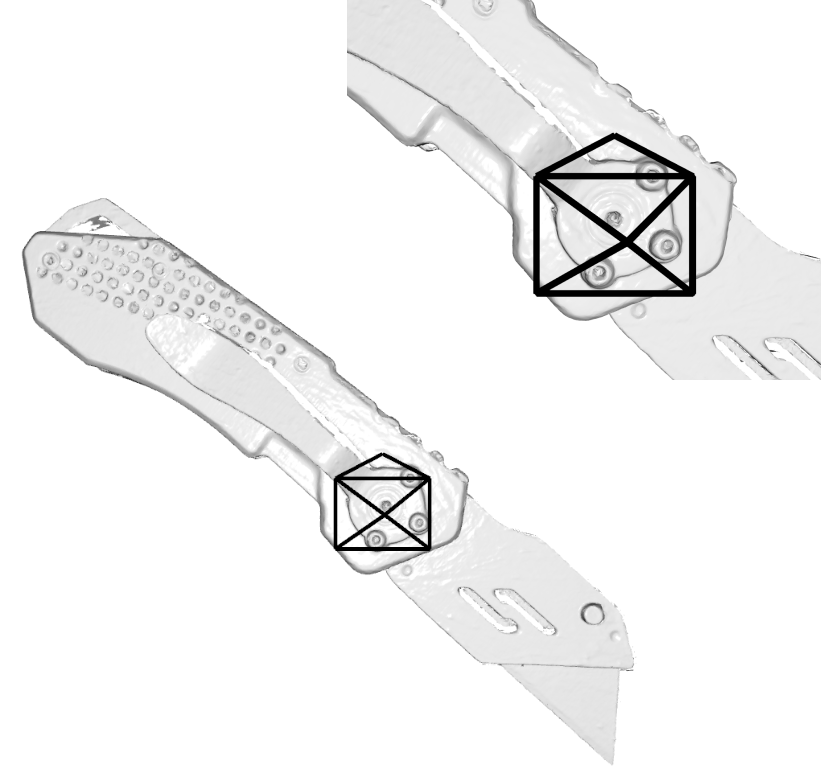}
  \caption{}
  \label{fig:pose_est_gs_final_folding_knife}
  \end{subfigure}
\caption{\Cref{fig:pose_est_rgb_folding_knife} is the pose estimated using the color channel image. This pose estimate was initialized with coarse pose estimates obtained using depth images. \Cref{fig:pose_est_gs_raw_folding_knife} is the GelSight data obtained at contact and \cref{fig:pose_est_gs_final_folding_knife} is the camera pose at contact. \Cref{fig:opening_figures,fig:pose_estimation_master} contain similar results.}
\label{fig:pose_estimation_folding_knife}
\end{figure}

Our RGBD sensor captures the depth of the scene reliably from 30 cm away and can produce aligned depth and color images. As the depth images are unchanged by changes in lighting and texture, extracting edges from depth images and maintaining coarse pose estimates as the robot moves closer to the object is a viable strategy to generate initial pose estimates which are later refined to estimate the pose at contact. For the objects described in \cref{fig:opening_figures,fig:pose_estimation_master,fig:pose_estimation_folding_knife}, when the depth sensor output degraded (at 20 cm) we transferred the pose estimates obtained (and maintained) using the depth image, to the color image.  refined the pose estimates from the final color image using the tactile image to obtain the pose at contact.  

\shrink
\subsubsection{Localizing small, flat and thin objects}\label{sc:loc_small_objects}
\begin{figure}[!]
    \begin{subfigure}[b]{0.15\textwidth}
        \centering
        \includegraphics[width=\textwidth]{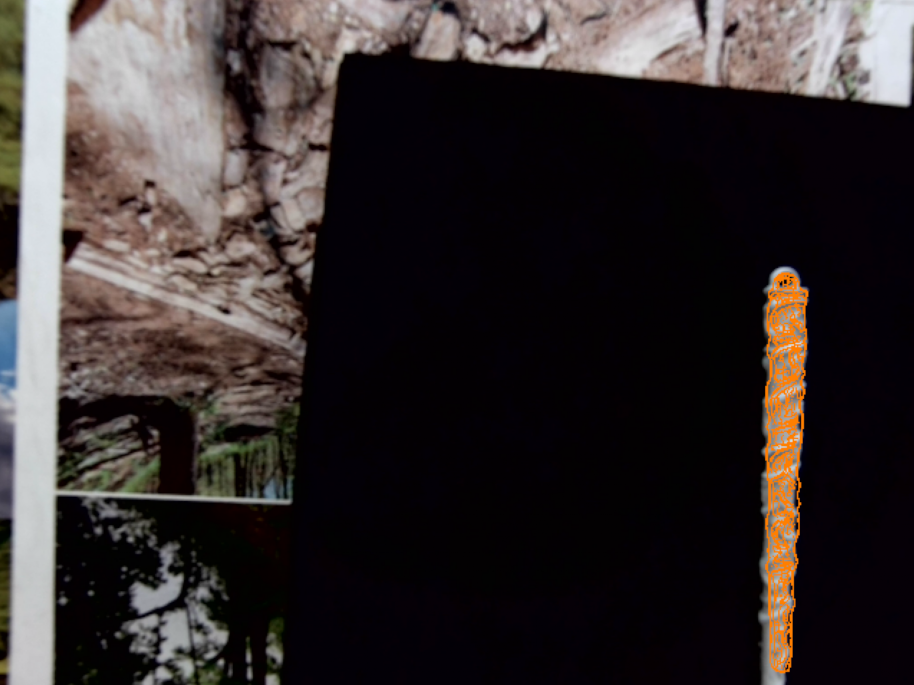}
        \caption{}
        \label{fig:pose_est_rgb_key_chain}
    \end{subfigure}
   \begin{subfigure}[b]{0.15\textwidth}
    \centering
    \includegraphics[width=\textwidth]{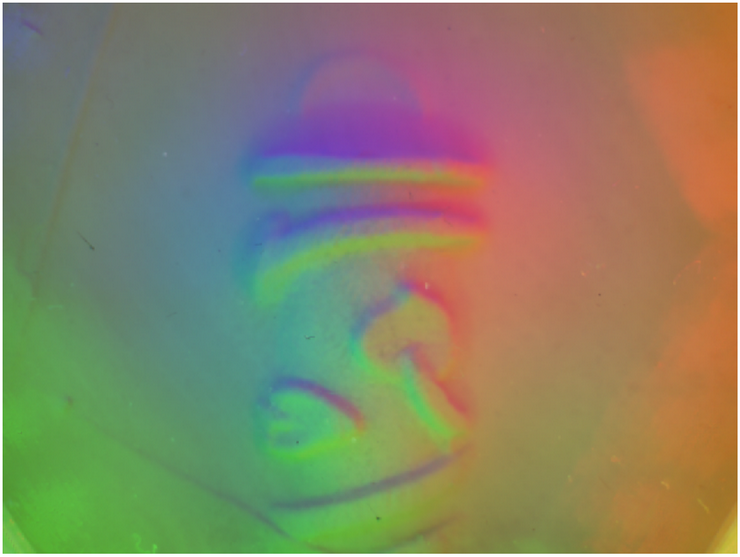}
    \caption{}
    \label{fig:pose_est_gs_raw_key_chain}
    \end{subfigure}
    \begin{subfigure}[b]{0.15\textwidth}
    \centering
    \includegraphics[width=0.6\textwidth]{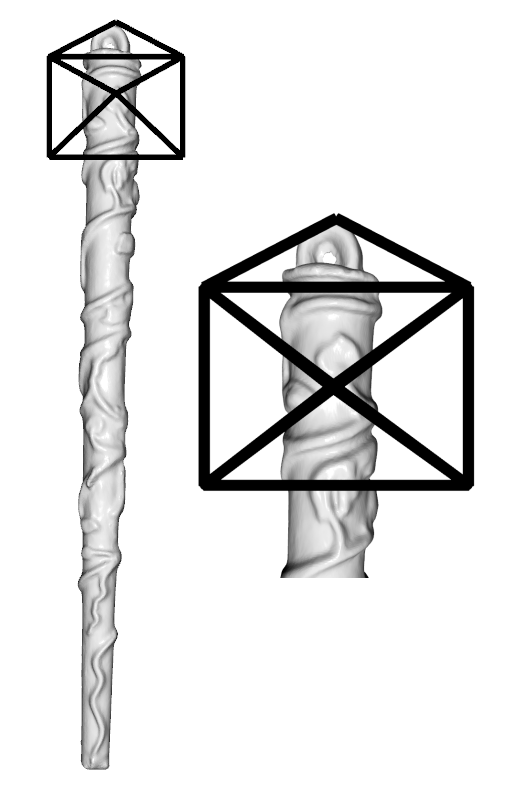}
    \caption{}
    \label{fig:pose_est_gs_final_key_chain}
    \end{subfigure}
\newline
    \begin{subfigure}[b]{0.15\textwidth}
    \centering
    \includegraphics[width=\textwidth]{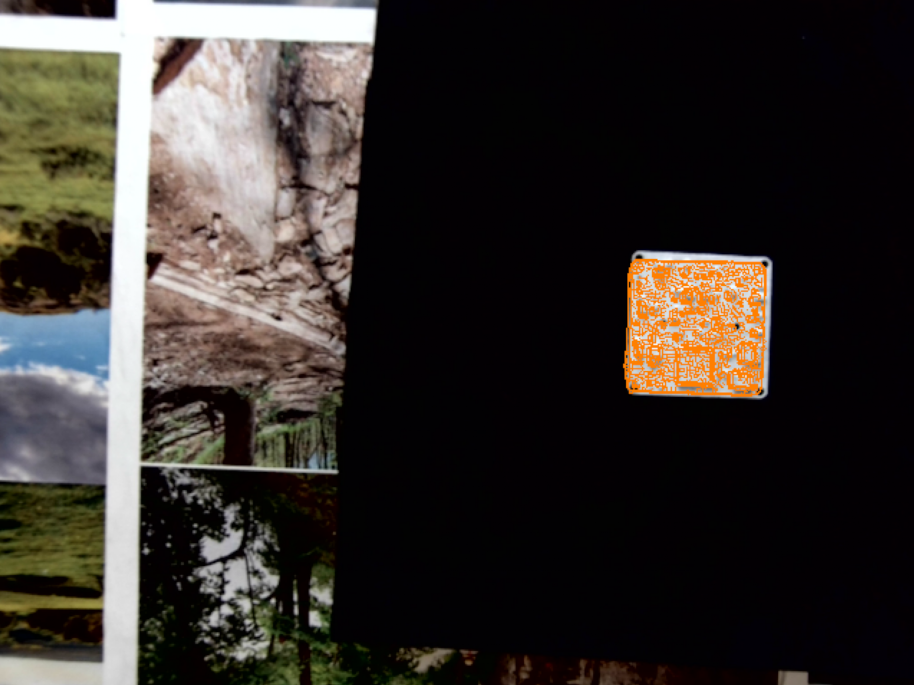}
    \caption{}
    \label{fig:pose_est_rgb_cam_ckt}
    \end{subfigure}
    \begin{subfigure}[b]{0.15\textwidth}
    \centering
    \includegraphics[width=\textwidth]{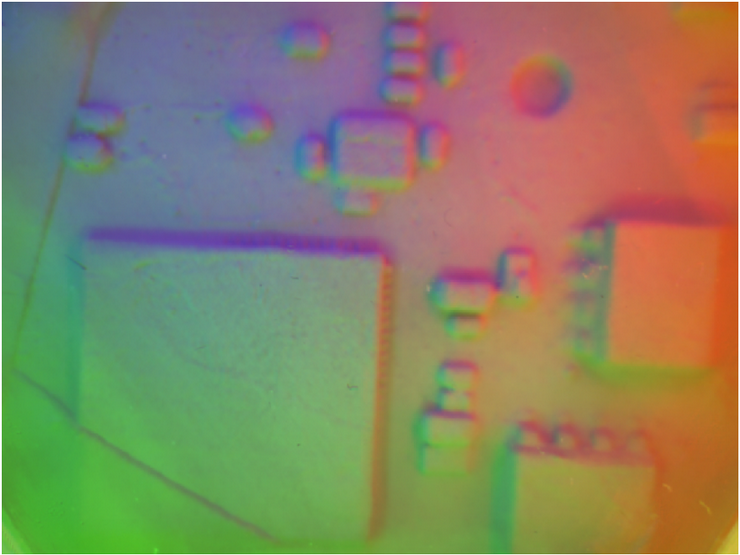}
    \caption{}
    \label{fig:pose_est_gs_raw_cam_ckt}
    \end{subfigure}
    \begin{subfigure}[b]{0.15\textwidth}
    \centering
    \includegraphics[width=0.8\textwidth]{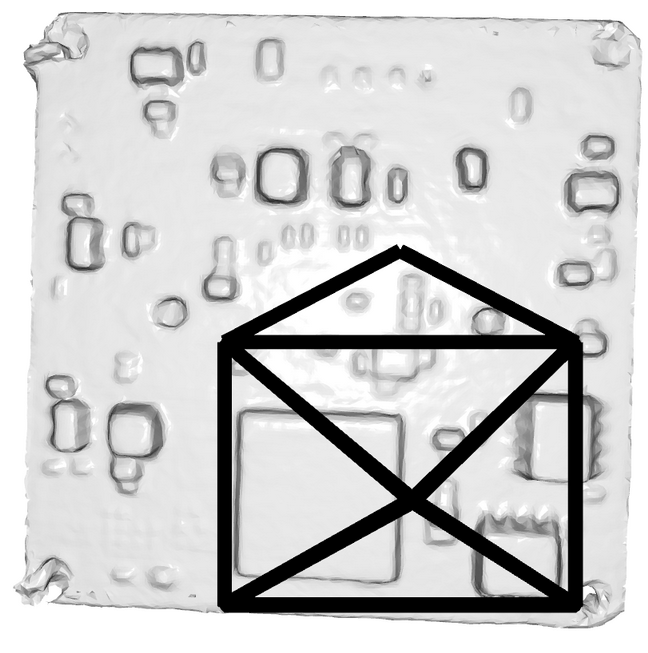}
    \caption{}
    \label{fig:pose_est_gs_final_cam_ckt}
    \end{subfigure}
    \caption{\Cref{fig:pose_est_rgb_key_chain,fig:pose_est_gs_raw_key_chain,fig:pose_est_gs_final_key_chain} are the steps in localizing a slender metallic object, \cref{fig:pose_est_rgb_cam_ckt,fig:pose_est_gs_raw_cam_ckt,fig:pose_est_gs_final_cam_ckt} are the steps in localizing a miniature circuit board. L-R we generate a coarse pose estimate with just the color image, receive the GelSight image on contact and finally obtain the camera pose on contact. The object in \cref{fig:pose_est_rgb_key_chain} is 10 cm long and 8.5 mm in diameter at the thickest part. The circuit board in \cref{fig:pose_est_rgb_cam_ckt} is 38mm $\times$ 38mm square. }

\label{fig:pose_estimation_key_chain_and_cam_ckt}
\end{figure}

Depth cameras don't capture much in situations where
an object has little depth variation. 
The limited field of view of most depth cameras means
they lose sight of where the GelSight sensor will
land as the hand approaches an object. For these reasons we
collocated a high resolution large field of view RGB camera with the GelSight sensor, so the system would
also work well with small thinner objects with less depth variation. In \cref{fig:pose_estimation_key_chain_and_cam_ckt} we demonstrate the localization of a slender metallic pin and
a 4cm square circuit board. 

\subsubsection{Disambiguating tactile measurements with vision}\label{sc:loc_disambiguate}
\begin{figure}[!]
    \begin{subfigure}[b]{0.15\textwidth}
        \centering
        \includegraphics[width=\textwidth]{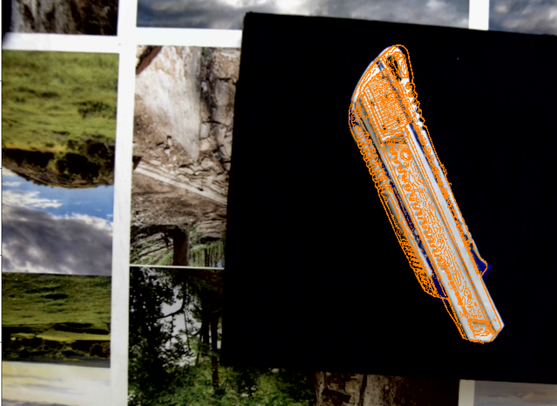}
        \caption{}
        \label{fig:pose_est_rgb_disambiguate}
    \end{subfigure}
   \begin{subfigure}[b]{0.15\textwidth}
    \centering
    \includegraphics[width=\textwidth]{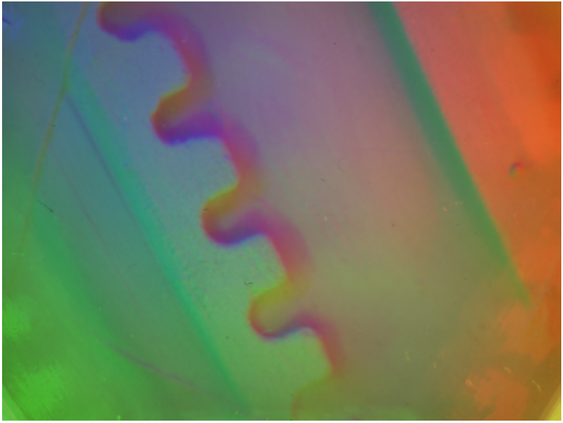}
    \caption{}
    \label{fig:pose_est_gs_raw_disambiguate}
    \end{subfigure}
    \begin{subfigure}[b]{0.15\textwidth}
    \centering
    \includegraphics[width=\textwidth]{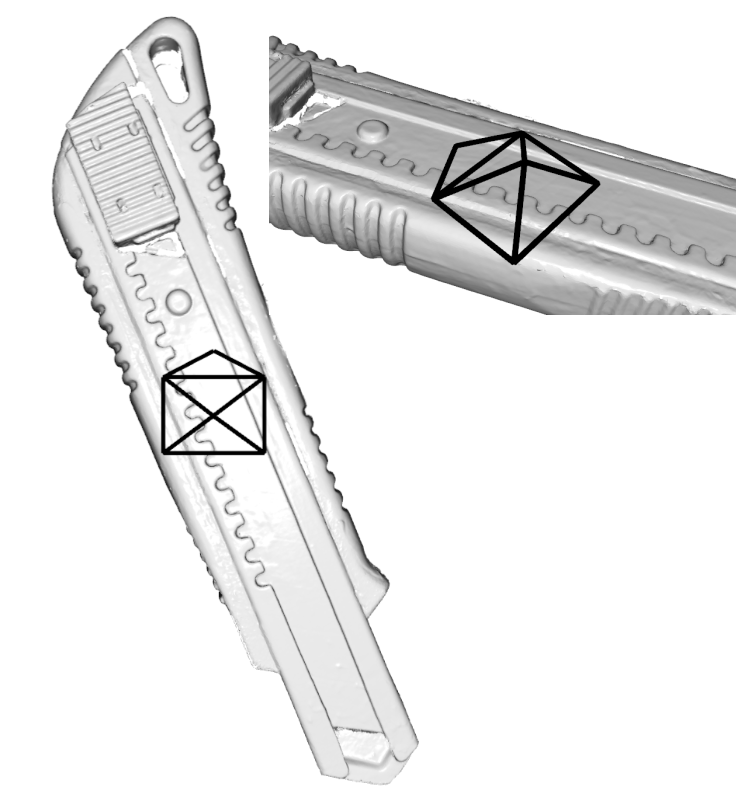}
    \caption{}
    \label{fig:pose_est_gs_final_disambiguate}
    \end{subfigure}
  \caption{Collocated vision can help us disambiguate between repeated tactile features. In this experiment we touch the middle of the set of teeth and could localize the contact using our pipeline.}
\label{fig:pose_estimation_disambiguate}
\end{figure}

With collocated vision, we can disambiguate between repetitive surface features, which would not have been possible with just tactile sensing. To show this, we set up an experiment (\cref{fig:pose_estimation_disambiguate}) where the robot approaches the box cutter from above and touches around the middle of the slider. The tactile image recorded is shown in \cref{fig:pose_est_gs_raw_disambiguate}. The pose estimates obtained from the camera when transferred to the GelSight yield useful initial guesses which can then be refined with the tactile data to localize the contact.  

\subsubsection{Estimating object pose with vision \textbf{and} touch}\label{sc:results_comparison}
\begin{table*}
   \begin{center}
    \begin{tabular}{|c||c|c|c|c|c|c||c|c|c|c|c|c|} 
      \hline
         Object (exp. name) & $\Delta T_x^v$ & $\Delta T_y^v$ & $\Delta T_z^v$ & $\Delta R_x^v$ & $\Delta R_y^v$ & $\Delta R_x^v$ & $\Delta T_x^t$ & $\Delta T_y^t$ & $\Delta T_z^t$ & $\Delta R_x^t$ & $\Delta R_y^t$ & $\Delta R_x^t$ \\
         \hline
         Box cutter slider (\ref{fig:pose_estimation_master}) & 2.5672  & 1.7059 & 5.4776 & 0.5199 & 0.3862 & 0.1761 & \cellcolor{green!20} 0.7174 & \cellcolor{green!20} 0.839 &  \cellcolor{green!20} 0.6012 & \cellcolor{green!20} 0.5031 & \cellcolor{green!20} 0.3692 & \cellcolor{green!20} 0.1842 \\
         \hline
         Box cutter teeth (\ref{fig:pose_estimation_disambiguate}) & 0.6949 & 0.5968 & 3.7251 & 0.131 & 0.1059 & 0.3534 &  \cellcolor{orange!20}6.7531 & \cellcolor{orange!20}1.9073 & \cellcolor{green!20}0.927 &  \cellcolor{orange!20}0.1433  & \cellcolor{green!20}0.0667 & \cellcolor{green!20}0.3014 \\
         \hline
         Folding knife (\ref{fig:pose_estimation_folding_knife})&  2.3934 & 3.4957& 8.4574& 0.3459& 0.2366 & 0.2701  &  \cellcolor{green!20}0.8343& \cellcolor{green!20}1.4932& \cellcolor{green!20}2.4594& 
\cellcolor{green!20}0.3543& \cellcolor{green!20}0.1062& \cellcolor{green!20}0.2392 \\
         \hline
         Glue Gun (\ref{fig:gelsight_pose_est}) &  1.0771& 1.0904& 8.3263& 0.1114& 0.1547& 0.2233  & \cellcolor{orange!20}2.1286 & \cellcolor{orange!20}2.2714 & \cellcolor{green!20} 1.8284&\cellcolor{green!20} 0.081 & \cellcolor{green!20}0.1175 &\cellcolor{green!20} 0.1628\\
         \hline
         Camera circuit (\ref{fig:pose_est_gs_final_cam_ckt}) &  0.2987 & 2.2541 & 2.2718  & 0.393 &  0.2401 & 2.1186 &  \cellcolor{green!20} 0.8923 &  \cellcolor{green!20} 0.3378 & \cellcolor{green!20} 1.0733  & \cellcolor{green!20} 0.3518 & \cellcolor{orange!20}0.2725 &  \cellcolor{green!20} 0.5157 \\
         \hline
         Metallic object (\ref{fig:pose_est_gs_final_key_chain}) &  0.6863 & 0.76 & 0.6261 & 0.0019 & 0.0012 & 0.0473 &  \cellcolor{green!20}0.3077 &  \cellcolor{orange!20}1.4643 & \cellcolor{green!20}0.0132 & \cellcolor{orange!20}0.002 & \cellcolor{orange!20}0.0057 & \cellcolor{green!20}0.0452 \\
         \hline
     \end{tabular}
     \caption{Relative uncertainties in localization using only vision and tactile sensing augmented by vision, in case good tactile signals are guaranteed. $\Delta T_x$ is the standard deviation in localization along the $x$ axis of the robot and $\Delta R_x$ is the standard deviation in orientation along the $x$ axis of the robot. Same naming convention applies to the $y$ and $z$ axes. The translation uncertainties are in mm. and the rotation uncertainties are in degrees.  Superscript $[\cdot]^v$ indicates localization using vision only and $[\cdot]^t$ indicates localization using vision and touch. Lower values are better in all cases. The position uncertainties are in mm and the angle uncertainties are in degrees.}
     \label{tab:col_vis_touch_expts}
   \end{center}
\end{table*}
In this section we present our results on localizing objects with vision and touch, and evaluate the performance of our localization pipeline. We focus on the case where the contact point has a unique arrangement of useful tactile features. This is the best case for a combined approach. In practice, if the tactile sensor contacts a featureless area, we can detect and ignore the output of the tactile sensor. Most work in this area specializes for a set of objects used in a training set for a learning approach. We did not find work that combined a camera with a high resolution tactile imaging sensor for localizing objects upon contact, so there isn't an obvious method to compare to. This along with our choice of smaller and highly textured objects preempts easy comparison with  recent learned localization approaches (e.g. \cite{Bauza2019, Bauza2020, Xiang2017} etc.). Instead, we present overall results of accuracy experiments below. \newline \indent For each of the 6 objects used in this work (\cref{fig:pose_estimation_master,fig:pose_estimation_disambiguate,fig:pose_est_gs_final_key_chain,fig:pose_estimation_folding_knife,fig:gelsight_pose_est,fig:pose_est_gs_final_cam_ckt}), we fix the object to the robot table. Assuming that there is zero error in position control of our robot (we use a Universal Robots UR5E manipulator fixed to a vention.io table of recommended design for the same robot), we register the object with respect to the robot base and treat this pose as our ground truth. Next, we move the robot vertically 1 m above the object and move the robot down to touch the object at a chosen point that will yield good tactile information, and localize the object with respect to the robot. We repeat this 3 times for the same object positions and repeat this experiment for 2 more positions of the object with respect to the robot -- i.e. localizing each object 9 times with respect to the robot. Fixing the objects is a restrictive assumption in the context of localizing objects especially with touch, however, to ensure repeatability of the experiments reported in the section, we had to fix the objects to a rigid base. Following \cite{Liu2012,Imperoli2015} we report the repeatibility of our pose estimation pipeline as the measure of its performance.  \newline
\indent  Using tactile sensing the localization errors were brought down to  $\pm1.5$mm in translation and $\pm0.5^\circ$ in rotation from about 1.5cm in translation and $2^\circ$ in rotation using only vision.  However, for cases where the tactile features were not unique, e.g. the box cutter teeth (\cref{fig:pose_estimation_disambiguate}) and the metallic object (\cref{fig:pose_est_gs_final_key_chain}), the tactile sensing actually increased the localization errors in the horizontal directions. The order of these errors were equivalent to the scale of the repeated features -- 5mm for the experiment described in \cref{fig:pose_est_gs_final_disambiguate} (the box-cutter teeth are about 3mm wide placed in intervals of 5mm) and about 2mm for the experiment described in \cref{fig:pose_est_gs_final_key_chain} (the embossed features are very similar at intervals of 3.5 mm). This observation is consistent with the fact that the final gradient descent step (\cref{eq:gelsight_alignment}) to refine the camera based pose estimates will converge to the wrong local minima if the tactile measurements are not distinctive enough. However, for objects with rich tactile features, using tactile sensing assisted with vision provides better localization than exclusively using either as we show in \cref{sc:exclusive_localization_vis}. We repeated a subset of the experiments reported above where we first corrected the robot trajectory using the procedure described in \cref{sc:Goal_correction} -- and observed similar localization performance. We present the results of the experiment reported in this section in \cref{tab:col_vis_touch_expts}.
\subsubsection{Effect of points of contact on localization accuracy}\label{sc:results_random_touch}
In this section we present the effect of randomly selected contacts for localization. For this set of experiments, we fix each of the objects and the black background plate used in \cref{sc:results_comparison} to a graduated compound slide capable of in plane translation and rotation. We then moved the robot vertically down to make contact at the same location on the object used for the experiments reported in \cref{sc:results_comparison} to generate a starting pose. Next, we generated 5 random configurations per object in translation and orientation on the plane of the table and moved the robot vertically down to touch the object and attempted to recover the randomly generated pose perturbations we introduced. For each of the objects, as expected, we observed similar errors in localization using only vision as reported in \cref{sc:results_comparison}. For the box cutter (\cref{fig:pose_estimation_master}) most of the contacts yielded useful tactile signals so the errors in recovering the perturbations in pose were in the range reported in \cref{sc:results_comparison} -- i.e. $\sim 8$mm in translation and $\sim 1.5^\circ$ in rotation. This observation was also consistent for the smaller textured objects$^3$ (\cref{fig:pose_estimation_key_chain_and_cam_ckt}). However, tactile sensing was not always helpful in localizing the objects -- for the glue gun (\cref{fig:opening_figures}) and the folding knife (\cref{fig:pose_estimation_folding_knife}, significant parts of the object were featureless and the tactile signals obtained when touched at these parts were unusable in localizing the objects as the final gradient descent step (\cref{eq:gelsight_alignment} in \cref{sc:gelsight_alignment}) re-introduced localization errors of about $3$-$4$ cm and $15^\circ$ by converging to incorrect poses. We report the results of the experiments described in this section in \cref{tab:vision_only_vs_vis_and_touch}.
\begin{table*}
    \centering
\begin{tabular}{|c|c|c|c|c|}
        \hline
        Object (exp. name)  & Ground Truth (x,y)  & Vision only (x,y) & Vision + Touch (x,y)  & Comment\\
        \hline
         Box cutter best case  &  41 (0, 41) & 50 (11, 49) & 45 (5, 45) & -- \\
        \hline
         Box cutter worst case  & 78 (0, 78)  & 89 (40, 79) & 55 (41, 35) & poor tactile signal due to interference\\
        \hline
         Metallic pin cutter best case  & 50 (0, 50)  & 40 (5, 40) & 53 (6, 52) & -- \\
        \hline
         Metallic pin cutter worst case  & 61 (0, 60) & 65 (8, 64) & 89 (10, 88) & partial contact between object and table \\
        \hline
         Glue gun best case  & 24 (24, 0) &  15 (12, 8) &  26 (25, 8) &  -- \\
        \hline
         Glue gun worst case  & 26 (24, 10) & 21 (20, 6) & 63 (60, 23) & tactile sensor touched flat part \\
        \hline
         Monkey best case & 13 (6, 12)  & 9 (7, 6) & 15 (8, 13) & -- \\
        \hline
         Monkey  worst case  & 14 (14, 4) & 18 (17, 6) & 36 (10, 35) & tactile sensor touched edge of arm and part of table \\
        \hline
    \end{tabular}
    \caption{Best and worst case localization errors when the contacts are selected at random. The ground truth measurements are calculated from the starting pose by reading the changes in X and Y translations on the graduated compound slide, the third column reports the magnitude of the translation recovered using vision only, and the fourth column reports the same using vision and touch. The X and Y components of the motion are indicated inside the brackets. We note that, even in the worst case, including the tactile measurements did not predict the contact in the wrong direction. In the last column, we provide possible explanations for the worst case errors as comments. All measurements have been rounded to mm. The errors in  orientation are within $2^\circ - 3^\circ$ for the cases when the pose estimation was successful and have not been reported. Some of the successful localization results can be viewed \href{https://arkadeepnc.github.io/projects/collocated_vision_touch/index.html}{on the paper website.} }
    \label{tab:vision_only_vs_vis_and_touch}
\end{table*}
\subsubsection{Localizing contact using only vision} \label{sc:exclusive_localization_vis}
% In the previous section we demonstrated our results of localizing objects with vision first and subsequently refining the localization estimates after receiving the information from the tactile sensor. 
 We explored two approaches for  predicting contact with only vision -- an image-based approach that tracks the POE similar to visual servoing and  a 3D geometric approach that
can estimate contact based on a single image captured before the object gets out of focus  or gets mostly out of the frame. \newline 
\indent In the \textbf{first case}, we identify the POE as described in \cref{sc:POE_description}.
Based on the known offset between the camera and the Gelsight, we can identify the predicted GelSight contact point in any
camera image. This approach works best when the hand is moving straight to the contact point (no trajectory corrections or going around obstacles), there are a lot of visual
features to track, and the hand starts high enough so there is little offset of the object and the background caused
by the different distances to the object and the background.
We tested performance with the data from the previous section for objects in \cref{fig:pose_estimation_master,fig:pose_estimation_disambiguate,fig:gelsight_pose_est}. We could predict the location of contact with 1cm accuracy.

In the \textbf{second case}, we took the final usable image in these trajectories and used the procedure described in \cref{sc:visual_pose_estimation} to obtain the pose of the object with respect to the RGB camera. Next, we transformed the pose of the object to the frame of the GelSight camera and simulated moving blindly until the vertical distance between the object and the GelSight is 3cm (the gel surface is 2.5 cm away from the GelGight camera), and reported the location of contact. We compared these estimates of the point of contact for all the objects we used. We found that for experiments with larger objects (as shown in \cref{fig:pose_estimation_master,fig:pose_estimation_disambiguate,fig:pose_estimation_folding_knife,fig:gelsight_pose_est}) the average error between the estimated true location of contact and the location estimated by blindly moving downwards in the frame of the GelSight camera was 3.5 cm, whereas the same metric for experiments with smaller objects (as shown in \cref{fig:pose_estimation_key_chain_and_cam_ckt}) were about 1.5 cm. The experiments with the smaller objects have lower errors because the distance between the GelSight and the object at the point the pose of the object was measured with only vision was much lower (8-10 cm) than the experiments with the larger object where the distance was about 25-30 cm, so the length of the blind descent was smaller for the smaller objects and hence the accumulated error in predicting the point of contact was smaller. \newline

\indent \textbf{To localize contact with only single instances of touch}, 
% \subsubsection{Localizing contact using only touch }\label{sc:exclusive_localization_touch}
To localize contact with only a single tactile image, we obtained a normal map and point cloud from the tactile image. We rendered a mesh model of the object (e.g. \cref{fig:pose_est_gs_final_disambiguate}) in a stable configuration, with the surface being touched facing the camera. The mesh is colored according to the normals with respect to a simulated camera, which is looking at the object vertically. Then we captured a high resolution image of the colored mesh and used ORB and SIFT features with FLANN based matching (ratio = 90\%) to find correspondences between our high resolution image of the rendered mesh and the normal map from the image. We also tried matching the raw images captured by the sensor (e.g. \cref{fig:pose_est_gs_raw_disambiguate}) with the corresponding image of the geometry (e.g. \cref{fig:pose_est_gs_final_disambiguate}). Finally, we rendered the mesh as a point cloud and used point cloud feature matching algorithms (B-SHOT\cite{Prakhya2017} and FPFH\cite{Rusu2009}) to generate correspondences between the template object and the tactile measurement processed as a point cloud, and identify the point of contact on the object. None of the above experiments were able to generate reliable correspondences between the object model and the tactile image. We observe that in the absence of good initial pose estimates from an external sensor or learned embeddings  (e.g. \cite{Bauza2020}), the portion of the object observed by the GelSight is often not large enough for conventional feature matching algorithms to work. Additionally, the GelSight elastomer mechanically operates as a nonlinear low pass spatial filter and does not preserve the true shape of depth discontinuities. Cliffs become gradual slopes, and the inverted surface
is filtered quite differently. As a result, point cloud feature matching often fails to match the sharp surface features from the object model to the data captured by the tactile sensor. The GelSight sensor elastomer mechanics need to be simulated to provide a better template to match. To localize contact with only a single tactile image, we attempted to match the processed tactile image obtained from the modified GelSight (\cref{fig:gelsight_cross_sn}) to a representation of the object model. First, we treated the tactile signal as an image (\cref{fig:gelsight_raw_data}) and extracted image features (SIFT and ORB) and attempted to match those features with a canonical image of the object mesh model. Next, we processed the raw GelSight data into a point cloud (\cref{fig:Gelsight_proc_data}) and used local point cloud feature matching algorithms (B-SHOT\cite{Prakhya2017} and FPFH\cite{Rusu2009}) between the processed tactile signal and the object mesh (e.g. \cref{fig:pose_est_gs_final}). None of the above experiments were able to generate reliable correspondences between the object model and the tactile image which leads us to conclude that in the absence of good initial pose estimates from an external sensor or learned embeddings  (e.g. \cite{Bauza2020}), the portion of the object observed by the GelSight is often not large enough for conventional feature matching algorithms to work.

%%%%%% 
\section{Discussion and future work}\label{sc:discussions}
%Why didn't we use a learning-based approach? 
We focused on objects with significant surface tactile texture that our small tactile sensor could image. This led us to not use existing sets of objects for manipulation research (YCB\cite{Calli2017}, McMaster\cite{Corona2018}). This different choice is due to the absence of a dataset to benchmark tactile imaging and vision working together. It is unclear what objects should be in the dataset, how approaches trained on such a dataset would generalize to novel objects and how much effort is needed to introduce new objects into the dataset and pipelines based on them. In our approach, we used an off the shelf 3D scanner (EinScan-SE\cite{einscan}) to generate 3D models of objects and reference poses for our initial estimates (see \cref{sc:visual_pose_estimation}).
% In our approach, where a model of the object can be created as the robot approaches an object, an object could be enrolled in the set of 3D models that provide reference poses for localization and pose estimation during the movement.
We do expect that versions of the current state-of-the-art edge based localization pipelines (what we use)  to be inherently slower than learned pipelines for generating initial pose estimates (e.g. \cite{Bauza2020} or \cite{Xiang2017}), but we believe that our localization pipeline would perform well for objects we have not tested here. Also, in the current literature, we did not find explicit or implicit feature transforms invariant across tactile images (or processed tactile data) and visual images. Although there exists work on learning features (see \cite{Bauza2019, Bauza2020, Li2019}), using them for explicit correspondences to match tactile data to visual data is relatively unexplored and we consider promising future work.  \newline
\indent 
While integrating  visual  sensors  of  different  capabilities,  we  noted that  having  optical sensors  with  almost  co-incident  optical axes  would  trivially  solve  some  of  the  issues we faced while localizing  small  objects  and  disambiguating  possible localizations. Collocating an ensemble of cameras of different capabilities (similar to smart phone multi-camera systems) creating a synthetic vision system is a natural next step of this work. Such a system could create a virtual fovea focused on where the tactile sensor could contact the object and bring down the overall footprint of the sensor setup so that it can be efficiently collocated with a standard gripper. Using a gripper with the ensemble of sensors is also future work. The GelSight data can be inexpensively processed to obtain surface normals of objects with respect to a fixed frame. This is valuable for pose estimation and should be emphasized in future object shape representations and manipulation algorithms.  Naturally, computing object surface normals by controlling illumination at the scale of the robot workspace is also future work. 

\section{Conclusion}
% In this work we show that collocating cameras with tactile sensors on a robot hand in many cases enables tactile localization to work, for the kinds of objects we tested and without substantial pre-computation using a known and fixed set of object models (learning). Vision  can also help disambiguate tactile image matching, especially in cases of limited, repetitive, or otherwise ambiguous tactile features. Tactile sensing almost always improves visual localization as well. We also found that using optic flow from hand mounted cameras had to be integrated across about 10cm of camera travel to provide useful heading estimates, and could be used to correct trajectories. 

In this work we show that collocating cameras with tactile sensors on a robot hand in many cases enables tactile localization to work where vision only or tactile sensing only localization may perform badly or fail. We demonstrated this with common objects without substantial pre-computation and without using a known and fixed set of object models (learning). Vision  can also help disambiguate tactile image matching, especially in cases of limited, repetitive, or otherwise ambiguous tactile features. Tactile sensing almost always improves visual localization as well. We also found that using optic flow from hand mounted cameras had to be integrated across about 10cm of camera travel to provide useful heading estimates, and could be used to correct trajectories. 

%%%%%%%%%%%%%%%%%%%%%%%%%%%%%%%%%%
\bibliographystyle{IEEEtran}
\bibliography{references}

\begin{thebibliography}{10}
\providecommand{\url}[1]{#1}
\csname url@samestyle\endcsname
\providecommand{\newblock}{\relax}
\providecommand{\bibinfo}[2]{#2}
\providecommand{\BIBentrySTDinterwordspacing}{\spaceskip=0pt\relax}
\providecommand{\BIBentryALTinterwordstretchfactor}{4}
\providecommand{\BIBentryALTinterwordspacing}{\spaceskip=\fontdimen2\font plus
\BIBentryALTinterwordstretchfactor\fontdimen3\font minus
  \fontdimen4\font\relax}
\providecommand{\BIBforeignlanguage}[2]{{%
\expandafter\ifx\csname l@#1\endcsname\relax
\typeout{** WARNING: IEEEtran.bst: No hyphenation pattern has been}%
\typeout{** loaded for the language `#1'. Using the pattern for}%
\typeout{** the default language instead.}%
\else
\language=\csname l@#1\endcsname
\fi
#2}}
\providecommand{\BIBdecl}{\relax}
\BIBdecl

\bibitem{Yamaguchi2016}
A.~Yamaguchi and C.~G. Atkeson,
  ``\href{https://ieeexplore.ieee.org/abstract/document/7803400}{Combining
  finger vision and optical tactile sensing: Reducing and handling errors while
  cutting vegetables},'' in \emph{2016 IEEE-RAS 16th International Conference
  on Humanoid Robots (Humanoids)}.\hskip 1em plus 0.5em minus 0.4em\relax IEEE,
  2016, pp. 1045--1051.

\bibitem{Yuan2017}
W.~Yuan, S.~Dong, and E.~H. Adelson,
  ``\href{https://www.mdpi.com/1424-8220/17/12/2762}{Gelsight: High-resolution
  robot tactile sensors for estimating geometry and force},'' \emph{Sensors},
  vol.~17, no.~12, p. 2762, 2017.

\bibitem{facebookresearch}
``\href{https://github.com/facebookresearch/Detectron}{{D}etectron: {FAIR}'s
  platform for Object Detection Research}.''

\bibitem{Song2020}
S.~Song, A.~Zeng, J.~Lee, and T.~Funkhouser,
  ``\href{https://arxiv.org/pdf/1912.04344.pdf}{Grasping in the wild: Learning
  6dof closed-loop grasping from low-cost demonstrations},'' \emph{IEEE
  Robotics and Automation Letters 2020}, vol.~5, no.~3.

\bibitem{Yamaguchi2017}
A.~Yamaguchi and C.~G. Atkeson,
  ``\href{https://ieeexplore.ieee.org/stamp/stamp.jsp?arnumber=8246881}{Implementing
  tactile behaviors using fingervision},'' in \emph{2017 IEEE-RAS 17th
  International Conference on Humanoid Robotics (Humanoids)}.\hskip 1em plus
  0.5em minus 0.4em\relax IEEE, 2017, pp. 241--248.

\bibitem{Yuan2016}
W.~Yuan, M.~A. Srinivasan, and E.~H. Adelson,
  ``\href{https://ieeexplore.ieee.org/document/7759057}{Estimating object
  hardness with a gelsight touch sensor},'' in \emph{2016 IEEE/RSJ
  International Conference on Intelligent Robots and Systems (IROS)}, pp.
  208--15.

\bibitem{Wang2018}
S.~Wang, J.~Wu, X.~Sun, W.~Yuan, W.~T. Freeman, J.~B. Tenenbaum, and E.~H.
  Adelson,
  ``\href{https://ieeexplore.ieee.org/stamp/stamp.jsparnumber=8593430}{3{D}
  shape perception from monocular vision, touch, and shape priors},'' in
  \emph{2018 IEEE/RSJ International Conference on Intelligent Robots and
  Systems (IROS)}.\hskip 1em plus 0.5em minus 0.4em\relax IEEE, 2018, pp.
  1606--13.

\bibitem{Smith2020}
E.~J. Smith, R.~Calandra, A.~Romero, G.~Gkioxari, D.~Meger, J.~Malik, and
  M.~Drozdzal, ``\href{https://arxiv.org/pdf/2007.03778.pdf}{3{D} shape
  reconstruction from vision and touch},'' \emph{arXiv preprint
  arXiv:2007.03778}, 2020.

\bibitem{Izatt2017}
G.~Izatt, G.~Mirano, E.~Adelson, and R.~Tedrake,
  ``\href{https://ieeexplore.ieee.org/stamp/stamp.jsp?tp=&arnumber=7989460}{Tracking
  objects with point clouds from vision and touch},'' in \emph{2017 IEEE
  International Conference on Robotics and Automation (ICRA)}.\hskip 1em plus
  0.5em minus 0.4em\relax IEEE, 2017.

\bibitem{Li2019}
Y.~Li, J.-Y. Zhu, R.~Tedrake, and A.~Torralba,
  ``\href{https://openaccess.thecvf.com/content_CVPR_2019/papers/Li_Connecting_Touch_and_Vision_via_Cross-Modal_Prediction_CVPR_2019_paper.pdf}{Connecting
  touch and vision via cross-modal prediction},'' in \emph{Proceedings of the
  IEEE/CVF Conference on Computer Vision and Pattern Recognition 2019}.

\bibitem{Luo2015_contact_localizaton}
S.~Luo, W.~Mou, K.~Althoefer, and H.~Liu,
  ``\href{https://ieeexplore.ieee.org/abstract/document/7139743}{Localizing the
  object contact through matching tactile features with visual map},'' in
  \emph{2015 IEEE International Conference on Robotics and Automation (ICRA)}.

\bibitem{Kelly2000}
R.~{Kelly}, R.~{Carelli}, O.~{Nasisi}, B.~{Kuchen}, and F.~{Reyes},
  ``\href{https://ieeexplore.ieee.org/stamp/stamp.jsp?tp=&arnumber=828588}{Stable
  visual servoing of camera-in-hand robotic systems},'' \emph{IEEE/ASME
  Transactions on Mechatronics}, vol.~5, no.~1, pp. 39--48, 2000.

\bibitem{Mori2013}
T.~Mori and S.~Scherer,
  ``\href{https://ieeexplore.ieee.org/stamp/stamp.jsp?tp=&arnumber=6630807}{First
  results in detecting and avoiding frontal obstacles from a monocular camera
  for micro unmanned aerial vehicles},'' in \emph{2013 IEEE International
  Conference on Robotics and Automation}.\hskip 1em plus 0.5em minus
  0.4em\relax IEEE, 2013, pp. 1750--1757.

\bibitem{Serres2017}
J.~R. Serres and F.~Ruffier,
  ``\href{https://www.sciencedirect.com/science/article/pii/S146780391730066X#cebib0010}{Optic
  flow-based collision-free strategies: From insects to robots},''
  \emph{Arthropod structure \& development}, vol.~46, no.~5, pp. 703--717,
  2017.

\bibitem{Raviv1992}
D.~Raviv,
  \emph{\href{https://nvlpubs.nist.gov/nistpubs/Legacy/IR/nistir4808.pdf}{A
  quantitative approach to looming}}.\hskip 1em plus 0.5em minus 0.4em\relax US
  Department of Commerce, National Institute of Standards and Technology, 1992.

\bibitem{Raviv2000}
D.~Raviv and K.~Joarder,
  ``\href{https://www.sciencedirect.com/science/article/abs/pii/S1077314200908622}{The
  visual looming navigation cue: a unified approach},'' \emph{Computer Vision
  and Image Understanding}, 2000.

\bibitem{Yang2020}
G.~Yang and D.~Ramanan,
  ``\href{https://openaccess.thecvf.com/content_CVPR_2020/papers/Yang_Upgrading_Optical_Flow_to_3D_Scene_Flow_Through_Optical_Expansion_CVPR_2020_paper.pdf}{Upgrading
  optical flow to 3d scene flow through optical expansion},'' in
  \emph{Proceedings of the IEEE/CVF Conference on Computer Vision and Pattern
  Recognition}, 2020.

\bibitem{Luo2015}
S.~Luo, W.~Mou, K.~Althoefer, and H.~Liu,
  ``\href{https://ieeexplore.ieee.org/stamp/stamp.jsp?arnumber=7109817}{Novel
  tactile-sift descriptor for object shape recognition},'' \emph{IEEE Sensors
  Journal}, vol.~15, no.~9, pp. 5001--5009, 2015.

\bibitem{Luo2016}
------,
  ``\href{https://ieeexplore.ieee.org/abstract/document/7759485}{Iterative
  closest labeled point for tactile object shape recognition},'' in \emph{2016
  IEEE/RSJ International Conference on Intelligent Robots and Systems
  (IROS)}.\hskip 1em plus 0.5em minus 0.4em\relax IEEE, 2016, pp. 3137--3142.

\bibitem{Donlon2018}
E.~Donlon, S.~Dong, M.~Liu, J.~Li, E.~Adelson, and A.~Rodriguez,
  ``\href{https://ieeexplore.ieee.org/abstract/document/8593661}{Gelslim: A
  high-resolution, compact, robust, and calibrated tactile-sensing finger},''
  in \emph{2018 IEEE/RSJ International Conference on Intelligent Robots and
  Systems (IROS)}.\hskip 1em plus 0.5em minus 0.4em\relax IEEE, 2018, pp.
  1927--1934.

\bibitem{Li2014}
R.~Li, R.~Platt, W.~Yuan, A.~ten Pas, N.~Roscup, M.~A. Srinivasan, and
  E.~Adelson,
  ``\href{https://ieeexplore.ieee.org/abstract/document/6943123}{Localization
  and manipulation of small parts using gelsight tactile sensing},'' in
  \emph{2014 IEEE/RSJ International Conference on Intelligent Robots and
  Systems}.\hskip 1em plus 0.5em minus 0.4em\relax IEEE, 2014, pp. 3988--3993.

\bibitem{She2019}
Y.~She, S.~Wang, S.~Dong, N.~Sunil, A.~Rodriguez, and E.~Adelson,
  ``\href{https://arxiv.org/abs/1910.02860}{Cable manipulation with a
  tactile-reactive gripper},'' 2019.

\bibitem{Bauza2019}
M.~Bauza, O.~Canal, and A.~Rodriguez,
  ``\href{https://ieeexplore.ieee.org/stamp/stamp.jsp?arnumber=8794298}{Tactile
  mapping and localization from high-resolution tactile imprints},'' in
  \emph{2019 International Conference on Robotics and Automation (ICRA)}.\hskip
  1em plus 0.5em minus 0.4em\relax IEEE, 2019.

\bibitem{Bauza2020}
M.~Bauza, E.~Valls, B.~Lim, T.~Sechopoulos, and A.~Rodriguez,
  ``\href{https://arxiv.org/pdf/2012.05205.pdf}{Tactile object pose estimation
  from the first touch with geometric contact rendering},'' \emph{arXiv
  preprint arXiv:2012.05205}, 2020.

\bibitem{Sodhi2020}
P.~Sodhi, M.~Kaess, M.~Mukadam, and S.~Anderson,
  ``\href{https://arxiv.org/pdf/2012.03768.pdf}{Learning Tactile Models for
  Factor Graph-based State Estimation},'' \emph{arXiv preprint
  arXiv:2012.03768}, 2020.

\bibitem{Alspach2019}
A.~Alspach, K.~Hashimoto, N.~Kuppuswamy, and R.~Tedrake,
  ``\href{https://ieeexplore.ieee.org/abstract/document/8722713}{Soft-bubble: A
  highly compliant dense geometry tactile sensor for robot manipulation},'' in
  \emph{2019 2nd IEEE International Conference on Soft Robotics
  (RoboSoft)}.\hskip 1em plus 0.5em minus 0.4em\relax IEEE, 2019, pp. 597--604.

\bibitem{Suresh2021}
S.~Suresh, Z.~Si, J.~G. Mangelson, W.~Yuan, and M.~Kaess,
  ``\href{https://arxiv.org/pdf/2109.09884.pdf}{Efficient shape mapping through
  dense touch and vision},'' \emph{arXiv preprint arXiv:2109.09884}, 2021.

\bibitem{Dikhale2022}
S.~Dikhale, K.~Patel, D.~Dhingra, I.~Naramura, A.~Hayashi, S.~Iba, and
  N.~Jamali, ``\href{https://ieeexplore.ieee.org/document/9682507}{VisuoTactile
  6D Pose Estimation of an In-Hand Object using Vision and Tactile Sensor
  Data},'' \emph{IEEE Robotics and Automation Letters}, 2022.

\bibitem{Xiang2017}
Y.~Xiang, T.~Schmidt, V.~Narayanan, and D.~Fox,
  ``\href{https://arxiv.org/abs/1711.00199}{Pose{CNN}: A convolutional neural
  network for 6d object pose estimation in cluttered scenes},'' \emph{arXiv
  preprint arXiv:1711.00199}, 2017.

\bibitem{Johnson2011}
M.~K. Johnson, F.~Cole, A.~Raj, and E.~H. Adelson,
  ``\href{https://dl.acm.org/doi/abs/10.1145/2010324.1964941}{Microgeometry
  capture using an elastomeric sensor},'' \emph{ACM Transactions on Graphics
  (TOG)}, vol.~30, no.~4, pp. 1--8, 2011.

\bibitem{OpenCV2021}
OpenCV,
  ``\href{https://docs.opencv.org/4.5.3/d2/d84/group__optflow.html#gad6aa63f2703202806fe18dc1353b5f4b}{Open
  Source Computer Vision Library},'' 2021.

\bibitem{Farnebaeck2003}
G.~Farneb{\"a}ck,
  ``\href{https://link.springer.com/chapter/10.1007/3-540-45103-X_50}{Two-frame
  motion estimation based on polynomial expansion},'' in \emph{Scand. conf. on
  Image anal.}\hskip 1em plus 0.5em minus 0.4em\relax Springer, 2003.

\bibitem{Brox2004}
T.~Brox, A.~Bruhn, N.~Papenberg, and J.~Weickert,
  ``\href{https://www.mia.uni-saarland.de/Publications/brox-eccv04-of.pdf}{High
  accuracy optical flow estimation based on a theory for warping},'' in
  \emph{European conference on computer vision}.\hskip 1em plus 0.5em minus
  0.4em\relax Springer, 2004, pp. 25--36.

\bibitem{Liu2012}
M.-Y. Liu, O.~Tuzel, A.~Veeraraghavan, Y.~Taguchi, T.~K. Marks, and
  R.~Chellappa,
  ``\href{https://journals.sagepub.com/doi/abs/10.1177/0278364911436018}{Fast
  object localization and pose estimation in heavy clutter for robotic bin
  picking},'' \emph{The International Journal of Robotics Research}, vol.~31,
  no.~8, pp. 951--973, 2012.

\bibitem{Imperoli2015}
M.~Imperoli and A.~Pretto,
  ``\href{https://link.springer.com/chapter/10.1007/978-3-319-20904-3_29}{$\mathrm{D^{2}CO}$:
  Fast and Robust Registration of 3D Textureless Objects Using the Directional
  Chamfer Distance},'' in \emph{International conference on computer vision
  systems}.\hskip 1em plus 0.5em minus 0.4em\relax Springer, 2015.

\bibitem{Choi2012}
C.~Choi and H.~I. Christensen,
  ``\href{https://ieeexplore.ieee.org/stamp/stamp.jsp?tp=&arnumber=6386065}{3d
  textureless object detection and tracking: An edge-based approach},'' in
  \emph{2012 IEEE/RSJ International Conference on Intelligent Robots and
  Systems}.\hskip 1em plus 0.5em minus 0.4em\relax IEEE, 2012.

\bibitem{Henderson2019}
P.~Henderson and V.~Ferrari,
  ``\href{https://link.springer.com/article/10.1007%2Fs11263-019-01219-8}{Learning
  single-image 3d reconstruction by generative modelling of shape, pose and
  shading},'' \emph{International Journal of Computer Vision}, pp. 1--20, 2019.

\bibitem{Felzenszwalb2012}
P.~F. Felzenszwalb and D.~P. Huttenlocher,
  ``\href{http://www.theoryofcomputing.org/articles/v008a019/v008a019.pdf}{Distance
  transforms of sampled functions},'' \emph{Theory of computing}, pp. 415--428,
  2012.

\bibitem{Prakhya2017}
S.~M. Prakhya, B.~Liu, W.~Lin, V.~Jakhetiya, and S.~C. Guntuku,
  ``\href{https://link.springer.com/article/10.1007/s10514-016-9612-y}{B-SHOT:
  a binary 3D feature descriptor for fast Keypoint matching on 3D point
  clouds},'' \emph{Autonomous Robots}, vol.~41, no.~7, 2017.

\bibitem{Rusu2009}
R.~B. Rusu, N.~Blodow, and M.~Beetz,
  ``\href{https://ieeexplore.ieee.org/stamp/stamp.jsp?arnumber=5152473}{Fast
  point feature histograms (FPFH) for 3D registration},'' in \emph{2009 IEEE
  international conference on robotics and automation}.\hskip 1em plus 0.5em
  minus 0.4em\relax IEEE, 2009, pp. 3212--3217.

\bibitem{Calli2017}
B.~Calli, A.~Singh, J.~Bruce, A.~Walsman, K.~Konolige, S.~Srinivasa, P.~Abbeel,
  and A.~M. Dollar,
  ``\href{https://journals.sagepub.com/doi/full/10.1177/0278364917700714}{Yale-CMU-Berkeley
  dataset for robotic manipulation research},'' \emph{The International Journal
  of Robotics Research}, vol.~36, no.~3, pp. 261--268, 2017.

\bibitem{Corona2018}
E.~Corona, K.~Kundu, and S.~Fidler,
  ``\href{http://www.cs.utoronto.ca/~ecorona/symmetry_pose_estimation/paper.pdf}{Pose
  estimation for objects with rotational symmetry},'' in \emph{2018 IEEE/RSJ
  International Conference on Intelligent Robots and Systems (IROS)}.\hskip 1em
  plus 0.5em minus 0.4em\relax IEEE, 2018, pp. 7215--7222.

\bibitem{einscan}
\BIBentryALTinterwordspacing
``Einscan-{SE}.'' [Online]. Available:
  \url{https://www.einscan.com/desktop-3d-scanners/einscan-se/}
\BIBentrySTDinterwordspacing

\end{thebibliography}

\end{document}